\documentclass[10pt,a4paper]{article}
\usepackage[english]{babel}
\usepackage[utf8]{inputenc}

\usepackage[calc]{datetime2}
\DTMnewdatestyle{versiondate}{%
}
\DTMnewtimestyle{versiontime}{%
}

\usepackage[
    a4paper,
    top=2cm,
    bottom=2cm,
    left=3cm,
    right=3cm,
    marginparwidth=1.75cm,
    includehead,
    headheight=15pt
]{geometry}

\usepackage{fancyhdr}
\usepackage{amsmath, amssymb, amsthm, mathtools}
\usepackage{enumitem}
\usepackage{booktabs}
\DeclareMathOperator*{\argmax}{arg\,max}

\numberwithin{equation}{section}
\newtheorem{theorem}{Theorem}[section]
\newtheorem{lemma}[theorem]{Lemma}
\newtheorem{remark}{Remark}

\newtheorem{proposition}{Proposition}
\newtheorem{definition}{Definition}
\newlist{assumptions}{enumerate}{1}
\setlist[assumptions]{
  label=\textbf{A\arabic*.},
  ref=A\arabic*
}
\usepackage{hyperref}
\usepackage[T1]{fontenc}
\usepackage{libertine}
\usepackage[libertine]{newtxmath}
\usepackage{inconsolata}
\usepackage{microtype}

\usepackage[
    backend=biber,
    style=alphabetic,
    url=false,
    isbn=false,
    doi=true,
    eprint=false,
    maxcitenames=2,
    mincitenames=1,
    maxbibnames=99
]{biblatex}
\AtEveryBibitem{
    \clearfield{urlyear}
    \clearfield{urlmonth}
}

\title{Goal Revelation as a Bridge Between \\
Information-Theoretic Empowerment and\\
Capability-Based Human Power}
\usepackage{authblk}

\author[1]{Jaime Ruiz-Serra}
\author[2]{Tom\'a\v{s} Gaven\v{c}iak}
\author[3]{Jobst Heitzig}
\affil[1]{Centre for Complex Systems, The University of Sydney}
\affil[2]{Alignment of Complex Systems Research Group, Center for Theoretical Study, Charles University, Prague}
\affil[3]{Complexity Science Department, Potsdam Institute for Climate Impact Research}

\usepackage{xcolor}
\definecolor{linksoft}{RGB}{70,95,150}
\definecolor{citesoft}{RGB}{45,120,105}
\hypersetup{colorlinks=true, linkcolor=linksoft, citecolor=citesoft, urlcolor=linksoft, filecolor=linksoft}

\newcommand{\empow}{\mathfrak{E}_h}
\newcommand{\Rev}{\mathrm{Rev}}
\newcommand{\Ch}{C_h}
\newcommand{\Cbar}{\bar{C}}
\newcommand{\Pbar}{\bar{P}}
\newcommand{\Bs}{B(s)}
\newcommand{\traj}{\xi}
\newcommand{\Trajs}{\Xi}
\newcommand{\Ngoals}{N}
\newcommand{\Stop}{\mathcal{S}^\top}
\newcommand{\Ghat}{\hat{G}}
\newcommand{\gnull}{g_0}
\newcommand{\pih}{\pi_h}
\newcommand{\pir}{\pi_r}
\newcommand{\pig}{\pi_g}
\newcommand{\Gh}{\mathcal{G}_h}
\newcommand{\Zpol}{Z}

\date{4 November 2026}

\begin{document}
\maketitle

\begin{abstract}
Power matters for AI safety.
Optimal agents tend to acquire power across a wide range of objectives (instrumental convergence).
Concurrently, as we embed AI agents into socio-technical systems, humans risk gradual disempowerment.
Both concerns call for a formal, quantitative measure of power.
Two such measures in the context of Markov games (multi-agent MDPs) are \emph{information-theoretic empowerment}, the capacity of the channel from an agent's actions to future states, and \emph{capability-based individual power}, a measure of goal attainment built from the probabilities that a human would attain each goal in a family of candidates were they pursuing it.
We show where the two meet. 
From individual power, Fano's inequality yields a floor $\Bs$, a strictly increasing function of individual power. 
This floor is a lower bound on what we term \emph{goal revelation}, the mutual information $\mathbb{I}(G;\traj \mid s)$ between the human's latent goal and the realised trajectory, or how much the interaction reveals about which goal was pursued. 
Empowerment caps revelation from above, through the directed information that the human's realised behaviour generates from actions to trajectory, so that $\Bs \le \mathbb{I}(G;\traj \mid s) \le \mathbb{I}(\mathbf{A}_h \to \traj \mid s) \le \empow(s)$.
Each quantity can be large while the one below it stays at zero, highlighting the following failure modes.
An assistant maximising empowerment can prefer an interface offering broad influence over goal-irrelevant degrees of freedom to one that lets the human attain every goal they might hold.
An assistant maximising revelation cannot tell a working interface from one rewired so that every goal yields the wrong outcome, since relabelling goals leaves mutual information unchanged while zeroing every attainment probability.
Positive guarantees therefore come only from the attainment end of the chain.
A decomposition traces unrevealed goals to either distinct goals inducing the same behaviour or behaviour unreadable from outcomes; a satisfaction floor, computable from two statistics of interaction logs, offers a way to monitor the power users retain in a deployed system; and simulations in random environments show the three quantities correlated but far from interchangeable.
\end{abstract}

\section{Introduction}
\label{sec:intro}

Power matters for AI safety on both sides of the human--AI relationship. Optimal agents tend to acquire power across a wide range of objectives \cite{Turner2021Power}, so recognising an AI that is amassing power requires the ability to measure power. And as AI agents are embedded into socio-technical systems, human control can erode through many individually unremarkable transfers \cite{Kulveit2025Gradual}, so avoiding gradual disempowerment requires tracking the power humans retain. Both concerns call for a formal, quantitative measure. Here we concern ourselves with two existing such measures. \emph{Empowerment} \cite{Klyubin2005EmpowermentUniversal} is the \emph{capacity} of the channel from an agent's actions to future states: a measure, in bits, of how much influence the environment affords them. \emph{Individual power} in the capability sense of \citeauthor{Heitzig2026FairObjective} \cite{Heitzig2026FairObjective} is a measure of goal \emph{attainment}: for each goal in a family of candidates, the (discounted) probability that an agent would attain it were they pursuing it, aggregated over the family. Whilst goals are a central object in the capability-based measure, they are completely absent in information-theoretic one. 

This paper connects the two through the mutual information $\mathbb{I}(G; \traj \mid s)$ between the goal $G$ of the agent of interest, treated as an unobserved random variable, and the state trajectory $\traj$ produced from the current state $s$. We call it \emph{goal revelation}, as it measures how much an observer of the trajectory could learn about which goal was being pursued. As we will show, it does not measure influence, since a agent whose goal-relevant options have all been removed cannot reveal information about the goal, however many other degrees of freedom they command. It also does not measure attainment, since a trajectory can identify the pursued goal while the goal goes unmet.

Our main result (Theorem~\ref{thm:bridge}) is a chain of bounds connecting the three quantities. From the individual power $I_h(s)$ (Equation~\ref{eq:individual-power}) proposed in~\cite{Heitzig2026FairObjective}, Fano's inequality yields a floor $\Bs$ \eqref{eq:fano-floor} on revelation. Revelation is capped in turn by the directed information $\mathbb{I}(\mathbf{A} \to \traj \mid s)$ that the realised behaviour generates, by Massey's converse for channels with feedback \cite{massey1990causality}, because the goal reaches the trajectory only through the actions. Finally, the directed information is capped by empowerment $\mathfrak{E}(s)$, because the realised behaviour is one feasible input among the ones the capacity maximises over. Together,
\[
\Bs \;\le\; \mathbb{I}(G;\traj \mid s) \;\le\; \mathbb{I}(\mathbf{A} \to \traj \mid s) \;\le\; \mathfrak{E}(s).
\]
The floor is a strictly increasing function of $I_h(s)$ that meets the maximum possible revelation at the maximum.

The converse implications all fail in ways that help us understand the problem. Capacity does not imply revelation: we exhibit environments in which the interface maximising $\empow$ offers unbounded capacity over degrees of freedom worthless to every candidate goal, while an alternative interface with maximal revelation scores arbitrarily worse (Proposition~\ref{prop:capacity-separation}). Revelation does not imply attainment: relabelling the goal--outcome interface by any permutation leaves $\mathbb{I}(G;\traj \mid s)$ unchanged while zeroing every attainment probability (Proposition~\ref{prop:scramble-separation}). Revelation is a property of the ensemble of goal-conditioned behaviours, indifferent to whether the goal each trajectory identifies is the goal it fulfils. Attainment capability restores the reference to fulfilment. Positive guarantees therefore come only from the attainment end of the chain.

Three further results sharpen the picture. We can decompose the induced behaviour class (Proposition~\ref{prop:z-decomposition}), splitting the observer's residual uncertainty into \emph{goal--policy aliasing}, when distinct goals induce the same behaviour, and \emph{policy--trajectory ambiguity}, when distinct behaviours produce indistinguishable outcomes. Revelation is thereby capped by the entropy of the behaviour ensemble, rather than the entropy of the goal ensemble. A satisfaction floor (Theorem~\ref{thm:vim}) extends the lower bound to families of overlapping goals. It is the divergence between the rate at which pursuit satisfies its goal and the rate at which chance would, two statistics estimable from interaction logs. Further, simulations in random one-step environments (Section~\ref{sec:experiment}) measuring how the three quantities co-vary in generic settings show that the ordering holds pointwise everywhere, and even in an ensemble that excludes both failure mechanisms by construction, the scores are correlated but far from interchangeable, agreeing on the best state about as often as a coin flip.

We make no claim that any of the three quantities is, by itself, a safe objective for an AI system. Here the aim is the relationship between the quantities, stated plainly and proved. In a companion manuscript we study{} what each quantity incentivises when maximised, including how an attainment capability objective can game the choice of its own goal family. 

\section{Setup}
\label{sec:setup}

For simplicity, our setup consists of a robot $r$ that wants to empower a single human $h$. We model the interaction as a fully observable, discrete-time Markov game (a multi-agent MDP). $\mathcal{S}$ is the finite state space, with current state $s \in \mathcal{S}$. At each step the agents choose actions $a_h \in \mathcal{A}_h$ and $a_r \in \mathcal{A}_r$ simultaneously, and the state moves by the kernel $\mathbb{P}(s' \mid s, a_h, a_r)$. Over a horizon $T$ we write $\mathbf{A}_h = (A_{h,1}, \dots, A_{h,T})$ for the human's action sequence viewed as a tuple of random variables.

The human holds a goal $g$ from a finite set $\Gh$ of $\Ngoals$ candidates, unobserved by the robot. $G$ denotes the goal viewed as a random variable. Section~\ref{sec:capability} provides a precise definition of a goal in terms of trajectory properties. Writing $\Delta(\mathcal{X})$ for the probability distributions over a finite set $\mathcal{X}$, the human acts by a goal-conditioned policy $\pih$ with $\pih(\cdot \mid s, g) \in \Delta(\mathcal{A}_h)$, which may also condition on the history of past states and own past actions where noted. The robot, with no access to $g$, acts by a state-conditioned Markov policy $\pir(\cdot \mid s) \in \Delta(\mathcal{A}_r)$. We write $\pig = (\pih(\cdot \mid \cdot, G{=}g), \pir)$ for the joint policy when the human pursues $g$.

A trajectory $\traj = (S_0, S_1, \dots, S_T)$ is the random state sequence generated by rolling out a joint policy from $S_0 = s$; every law written $(\cdot \mid s)$ is the law of this rollout. All per-interaction quantities are evaluated over a finite window of $T$ steps, making the trajectory space is $\Trajs \coloneqq \mathcal{S}^{T+1}$. An environment whose interaction ends early, in an absorbing state, is read as padded to the full window by repetition of that state. Interaction is segmented into such windows, and the goal is fixed within a window. The distribution the goal is drawn from, and its interpretation, are discussed in Assumption~\ref{ass:goal-prior}, Section~\ref{sec:revelation}. Throughout, logarithms are base 2 and information is measured in bits. $\mathbb{H}$ is entropy, $\mathbb{I}$ mutual information (directed where marked with an arrow, introduced in Section~\ref{sec:empowerment}), and $\mathbf{1}(\cdot)$ the indicator function.

\section{Three Measures}
\label{sec:measures}

\subsection{Empowerment: Potential Influence}
\label{sec:empowerment}

Empowerment \cite{Klyubin2005EmpowermentUniversal} is the capacity of the channel from an agent's actions to future states. Klyubin et al.\ defined it as the $n$-step open-loop capacity between an action sequence and the endpoint state $S_n$. Since the human of Section~\ref{sec:setup} observes the state as they act, we use the closed-loop form on the full trajectory (in the spirit of \citeauthor{Tiomkin2025ProcessEmpowerment} \cite{Tiomkin2025ProcessEmpowerment}) defining the capacity in Massey's directed information \cite{massey1990causality},
\begin{equation}
    \empow(s \mid \pir)
    \coloneqq \max_{p(\mathbf{a}_h \parallel \traj_{0:T-1})} \mathbb{I}_{\pir}(\mathbf{A}_h \to \traj \mid s)
    = \max_{p(\mathbf{a}_h \parallel \traj_{0:T-1})} \sum_{t=1}^T \mathbb{I}_{\pir}(\mathbf{A}_{h,1:t} \,;\, S_t \mid \traj_{0:t-1})
    \;\;\le\;\; T\log_2|\mathcal{A}_h|.
    \label{eq:empowerment}
\end{equation}
The maximisation ranges over all history-dependent behaviour policies for the human, defined as causally conditioned input distributions $p(\mathbf{a}_h \parallel \traj_{0:T-1}) = \prod_{t=1}^T p(a_{h,t} \mid \mathbf{a}_{h,1:t-1}, \traj_{0:t-1})$, in which each action may depend on everything realised so far but not subsequent states. This maximiser is a device for scoring the state $s$, not a model of what the human will do. In particular, it is not the goal-conditioned policy $\pih$ (and no goal variable appears in Equation~\eqref{eq:empowerment}). The channel being maximised against is the environment with the robot's action marginalised out,
\begin{equation}\label{eq:induced-channel}
    p_{\pir}(s_t \mid s_{t-1}, a_{h,t}) = \sum_{a_r} \pir(a_r \mid s_{t-1})\; \mathbb{P}(s_t \mid s_{t-1}, a_{h,t}, a_r),
\end{equation}
extended over the window as a controlled Markov chain. The feasible set of input distributions is a compact convex polytope and directed information is continuous on it, so the maximum is attained. We adopt the following two conventions, which will matter for comparisons later on. $\empow(s \mid \pir)$ is an unnormalised $T$-use capacity, in bits per window rather than per step, in order to make it comparable with the goal entropy $\log_2 \Ngoals$. The closed-loop form dominates the open-loop and endpoint-state forms: open-loop action sequences are among the policies maximised over, for them directed information equals mutual information \cite{massey1990causality}, and the endpoint state is part of the trajectory. It is therefore the most generous choice of ``influence afforded'' quantity\footnote{The separation of Section~\ref{sec:separations} is not an artefact of the reformulation, since in its deterministic instance, the closed- and open-loop values coincide.}. 
As with any capacity, the operational interpretation is a coding statement over repeated uses (in the limit). In the deterministic environments of our worked instances it reduces to a zero-error count, with $\empow(s\mid\pir)$ equal to $\log_2$ of the number of distinguishable trajectories\footnote{We evaluate all objectives at the current state, as an intrinsic reward would be. A deployed objective would aggregate over the states the robot's policy visits, which can change the policy's score. The comparisons in this paper are between scores at one fixed $s$.}.

\subsection{Capability: Attainment}
\label{sec:capability}

\citeauthor{Heitzig2026FairObjective} \cite{Heitzig2026FairObjective} measure power by attainment instead. For a state $s$ and a candidate goal $g$, the question is counterfactual: how capable of attaining $g$ is the human, were they pursuing $g$? In their setting a goal is a proposition about the world state, represented by the set of states $\mathcal{S}_g \subseteq \mathcal{S}$ in which it holds. The game form is finite and acyclic with a terminal set $\Stop$. The human's behaviour is a given goal-conditioned policy that is the robot's model of the human, rather than an optimality assumption. The capability is determined by six axioms on how attainment may depend on the trajectory law, up to a monotone transform, yielding the truncated Bellman recursion
\begin{equation}
    \Ch(s, g \mid \pig) = \mathbf{1}(s \in \mathcal{S}_g) + \mathbf{1}(s \notin \mathcal{S}_g \cup \Stop)\; \gamma_h \, \mathbb{E}_{s' \sim \pig, s} \big[\Ch(s', g \mid \pig)\big] \;\in\; [0,1],
    \label{eq:capability-recursion}
\end{equation}
with discount $\gamma_h \in (0,1)$, as the simplest admissible choice. This is the discounted probability that pursuit of $g$ reaches it. We adopt the recursion as it stands. It is evaluated under the goal-conditioned policy $\pig$ of Section~\ref{sec:setup}, the robot's model of the human\footnote{Every result below holds for whichever policy is used. Under a model it bounds the revelation of the modelled human, and carries over to the actual human as far as the model is accurate (Section~\ref{sec:discussion}).}. We depart from their setting in two places. They impose \emph{coverage}, meaning every state lies in some goal, so the current state always carries one attained goal and individual power below is non-negative. We adopt the opposite, i.e. no goal is certified at time zero (Assumption~\ref{ass:atomic-goals}), since a window that decodes before anyone acts is trivial. The goal family need not cover every trajectory, since a window may end in $\Stop$, or run out, with no goal satisfied. Every capability can then vanish and $I_h$ diverge to $-\infty$, which the floor of Theorem~\ref{thm:bridge} tolerates because it clamps at zero. We also admit the endpoint $\gamma_h = 1$, since the finite window needs no discount for the recursion to be well posed. The capability is then the plain probability of reaching $\mathcal{S}_g$ before $\Stop$, the setting of our worked instances.

``Eat an apple after finishing the coffee'' and ``never enter the room'' constrain the path rather than a single state, so neither is a set of states in the state space as modelled. Both become sets of states once a state records the history that reached it, and the recursion above then applies unchanged\footnote{Unroll the window into its prefix tree: a state is a prefix $\traj_{0:t}$ and the terminal states are the full trajectories. The state set induced by a trajectory goal $\Trajs_g$ collects the prefixes all of whose extensions satisfy $g$. First entry into it is the certificate time of Definition~\ref{def:trajectory-goal}, and Equation~\eqref{eq:capability-recursion} on the tree returns the capability of Equation~\eqref{eq:capability-def}. Appendix E.2 of \cite{Heitzig2026FairObjective} reads goals this way for non-tree transition graphs, as finite continuations of the trajectory from $s$. Two caveats. Extensions are taken in $\Trajs = \mathcal{S}^{T+1}$, so a prefix certifies only when every combinatorial extension satisfies $g$. The same construction on the reachable tree certifies no later, and at $\gamma_h < 1$ sometimes strictly earlier, which makes ours the conservative reading. At $\gamma_h = 1$, the value used in every instance here, the two agree, both being the satisfaction probability. The tree does not reconcile the frameworks either: its root is a state, so coverage would demand a goal certified there, which Assumption~\ref{ass:atomic-goals} excludes.}. We work with the trajectory property directly. It gives the certificate time an explicit form, and it keeps goals in the same space as the other two measures.

\begin{definition}[Trajectory goals and certificate times]
\label{def:trajectory-goal}
A \emph{goal} $g$ is identified with a nonempty \emph{satisfaction set} $\Trajs_g \subseteq \Trajs$, such that $\traj \in \Trajs_g$ means ``$\traj$ satisfies $g$''. The \emph{certificate time} of $g$ is the time index
\begin{equation}
    \tau_g(\traj) \coloneqq \min\big\{ t \ge 0 : [\traj_{0:t}] \subseteq \Trajs_g \big\}, \qquad \min\emptyset \coloneqq \infty,
    \label{eq:certificate-time}
\end{equation}
where $[\traj_{0:t}]$ collects the trajectories sharing the prefix $\traj_{0:t}$: the first step at which the prefix alone guarantees satisfaction.
\end{definition}

A prefix with $[\traj_{0:t}] \subseteq \Trajs_g$ is a good prefix in the sense of \citeauthor{KupfermanVardi2001Safety} \cite{KupfermanVardi2001Safety}. The definition also sorts goals along the safety--co-safety axis \cite{MannaPnueli1990Hierarchy}: reaching goals are certified at first entry, while ``never enter the room'' is certified only at the window's end, $\tau_g = T$. Capability generalises as the expected discounted certificate,
\begin{equation}
    \Ch(s, g \mid \pig) \coloneqq \mathbb{E}_{\pig, s}\big[\gamma_h^{\tau_g(\traj)}\,\mathbf{1}(\traj \in \Trajs_g)\big], \qquad \gamma_h^{\infty} \coloneqq 0.
    \label{eq:capability-def}
\end{equation}
which is the satisfaction probability when $\gamma_h = 1$, and which additionally accounts for how soon satisfaction is guaranteed when $\gamma_h < 1$. For the state-set goal ``reach $\mathcal{S}_g$ before $\Stop$'', the certificate time is the first entry into $\mathcal{S}_g$ and Equation~\eqref{eq:capability-def} reduces to the recursion~\eqref{eq:capability-recursion}.

A single trajectory can visit several target sets in sequence. From pairwise-disjoint target sets $\{\mathcal{S}_g\}_g$, the \emph{first-outcome goal} of $\mathcal{S}_g$ has satisfaction set
\begin{equation}
    \Trajs_g \coloneqq \big\{ \traj \in \Trajs : \traj \text{ enters } \mathcal{S}_g \text{ at some } t \le T \text{ without previously entering } \textstyle\bigcup_{g' \ne g} \mathcal{S}_{g'} \cup \Stop \big\},
    \label{eq:first-outcome-atom}
\end{equation}
whose members are the trajectories in which $g$ is the \emph{first} outcome realised. 
These satisfaction sets are pairwise disjoint by construction. 
First-outcome goals are an analysis device, not a claim that humans care about outcome order. 
Disjointness makes goal identity readable off a trajectory (Section~\ref{sec:bridge}). 
Under this framing, attainment via a detour through another target counts as failure, which loosens the bounds. 
Section~\ref{sec:overlap} provides an alternative framing for instances where goal regions interleave. 
Equation~\eqref{eq:capability-def} evaluates first-outcome goals directly, since its certificate time is the first entry into $\mathcal{S}_g$, before any other target and any state of $\Stop$. For such a goal, $\Ch(s, g \mid \pig)$ is the discounted probability that pursuing $g$ makes it the window's first outcome.

Individual power aggregates the capabilities over the candidate goal family with an exponent $\zeta$, without asking which goal the human holds. \citeauthor{Heitzig2026FairObjective} derive its form from axioms on the aggregate, among them neutrality to goal labels, invariance to adding unreachable goals, a preference for reliable over uncertain attainment, and additivity over independent games:
\begin{equation}
    I_h(s \mid \pir) \coloneqq \log_2 \sum_{g \in \Gh} \Ch(s, g \mid \pig)^{\zeta}.
    \label{eq:individual-power}
\end{equation}
Their reliability axiom forces $\zeta > 1$. We allow any $\zeta \ge 1$, which Theorem~\ref{thm:bridge} covers. Small $\zeta$ favours breadth (many moderate capabilities), large $\zeta$ favours peak capability. Each summand conditions the human's policy on a different hypothetical goal while the robot's policy stays fixed and goal-agnostic. The aggregate has a Bayesian reading that will pair with the observer of the next subsection: taking $p(\mathcal{O} = 1 \mid s, g) \coloneqq \Ch(s, g \mid \pig)^\zeta$ as the likelihood of an abstract success event $\mathcal{O}$ under a uniform prior over goals, $I_h(s) = \log_2 \Ngoals + \log_2 p(\mathcal{O} = 1 \mid s)$ is the log-evidence of the model in which the human is observed to succeed, and the associated posterior over goals is $\Ch^\zeta(s, g \mid \pi_g) / \sum_{g'} \Ch^{\zeta}(s, g' \mid \pi_{g'})$. Individual power thus scores $s$ by how credible success is from there, one hypothetical goal at a time.

\subsection{Goal Revelation: Identifiability}
\label{sec:revelation}

The third quantity needs the goal as a random variable with the following assumption.

\begin{assumptions}
    \item \label{ass:goal-prior} \textbf{Goal prior:} The goal is fixed for the window and unobserved. We model it as drawn before the window begins from a reference prior $p(g)$ over $\Gh$, uniform unless stated otherwise, $p(g) = 1/\Ngoals$. The prior belongs to whoever would infer the goal from the trajectory, not to whoever scores the state. It captures a designer's or observer's ignorance of which goal this human holds, or an empirical frequency over a population of interaction windows where one exists (as in the deployment logs of Section~\ref{sec:overlap}). It is fixed before interaction and does not depend on the robot's policy. Individual power~\eqref{eq:individual-power}, the score the robot itself computes, carries no prior: the robot knows the family $\Gh$ but holds no belief over which member is live. Every information quantity involving $G$ is relative to the prior.
\end{assumptions}

\begin{definition}[Goal revelation]
\label{def:revelation}
Given policies $(\pih, \pir)$ and the prior of Assumption~\ref{ass:goal-prior}, \emph{goal revelation} at $s$ is the mutual information between goal and trajectory,
\begin{equation}
    \Rev(s \mid \pih, \pir) \coloneqq \mathbb{I}_{\pih,\pir}(G; \traj \mid s) = \mathbb{H}(G \mid s) - \mathbb{H}(G \mid \traj, s).
    \label{eq:revelation}
\end{equation}
\end{definition}

We write $\Rev(s)$, and likewise $\empow(s)$ and $I_h(s)$, when the policies are fixed and understood. Revelation is what an observer of the interaction could learn: watching the trajectory and updating the prior to the posterior $p(g \mid \traj, s)$, they gain $\Rev(s)$ bits about the pursued goal on average. Equivalently, by \eqref{eq:revelation}, revelation is the prior entropy minus the \emph{equivocation} $\mathbb{H}(G \mid \traj, s)$, the uncertainty that remains after the trajectory has been seen. It is capped by the ceiling $\Rev(s) \le \mathbb{H}(G \mid s) = \log_2 \Ngoals$, however rich the environment.

Revelation, as defined, has three important properties. 
\begin{itemize}
    \item It is zero when behaviour is not goal-conditioned. Under a policy that ignores $g$, the joint law factorises and $\mathbb{I}(G;\traj \mid s) = 0$. Acting differently for different goals is the only channel through which the goal reaches the world.
    
    \item It is a property of the \emph{ensemble} of goal-conditioned behaviours (rather than of any single window). If each goal induces a deterministic policy with a distinct outcome, then every window is deterministic given the goal and $\Rev(s) = \mathbb{H}(G \mid s)$ is maximal purely because the trajectory identifies which policy the ensemble selected.
    
    \item It is indifferent to attainment. The definition does not consider whether $\traj$ satisfies the goal that produced it, only whether trajectories discern goals, which is why we call it revelation and not control. Section~\ref{sec:separations} provides a theorem concerning this issue.
\end{itemize}
Prior work has optimised the same functional with the latent variable owned by the agent itself, as skill discovery. In Section~\ref{sec:related} we discuss how, in the skill discovery framing, relabelling the latent is a symmetry rather than a failure.

\section{The Bridge}
\label{sec:bridge}

The first-outcome construction of Section~\ref{sec:capability} helps us construct a further structural assumption needed for decoding.

\begin{assumptions}[resume]
    \item \label{ass:atomic-goals} \textbf{Disjoint goals:} The satisfaction sets $\{\Trajs_g\}_{g \in \Gh}$ are pairwise disjoint (no trajectory satisfies two goals). The family need not be exhaustive (a trajectory may not satisfy any goal). None of the goals are certified at time zero from the initial state ($\tau_g \ge 1$ for every $g \in \mathcal{G}_h$), which excludes windows that decode trivially before anyone acts.
\end{assumptions}

\begin{theorem}[Capacity, revelation, attainment]
\label{thm:bridge}
Fix $s$, policies $(\pih, \pir)$ with $\pir$ Markov, the prior of Assumption~\ref{ass:goal-prior}, and a horizon $T$. Then
\begin{equation}
    \underbrace{\Bs}_{\textnormal{floor from attainment}} \;\le\; \underbrace{\Rev(s)}_{\textnormal{revelation}} \;\le\; \underbrace{\mathbb{I}_{\pih,\pir}(\mathbf{A}_h \to \traj \mid s)}_{\textnormal{realised influence-information}} \;\le\; \underbrace{\empow(s)}_{\textnormal{capacity}},
    \label{eq:bridge-chain}
\end{equation}
where the middle quantity is the directed information generated by the human's realised, goal-marginalised behaviour, and the floor, which additionally requires the uniform prior (\ref{ass:goal-prior}), disjoint goals (\ref{ass:atomic-goals}), and capabilities per Equation~\eqref{eq:capability-def} with any $\gamma_h \le 1$, $\zeta \ge 1$, is
\begin{equation}
    \Bs \coloneqq \max\Big(0,\; \log_2 \Ngoals - f\big(\min\big(P_e^{\max}(s), \tfrac{\Ngoals - 1}{\Ngoals}\big)\big)\Big),
    \qquad P_e^{\max}(s) = 1 - 2^{\,I_h(s) - \log_2 \Ngoals},
    \label{eq:fano-floor}
\end{equation}
with $f(x) = h_2(x) + x\log_2(\Ngoals - 1)$ and $h_2$ the binary entropy function.
\end{theorem}

The full proof is in Appendix~\ref{app:proofs}. The floor uses a decoding argument: an observer who watches the trajectory guesses the goal by satisfaction, $\Ghat = g$ if and only if $\traj \in \Trajs_g$, single-valued by disjointness and defaulting to a fixed $\gnull$ on trajectories that do not satisfy a goal. Since $\gamma_h \le 1$ and $x^\zeta \le x$ on $[0,1]$, the average correct-guess probability is at least $2^{I_h(s) - \log_2 \Ngoals}$ and we obtain the resulting error bound~\eqref{eq:fano-floor} from Fano's inequality. The clamp at the random-guessing rate $(\Ngoals-1)/\Ngoals$ prevents the bound from asserting a positive floor even when every capability is zero ($I_h \to -\infty$, $P_e^{\max} \to 1$). The middle inequality is Massey's converse for channels with feedback \cite{massey1990causality}: the kernel~\eqref{eq:induced-channel} does not consider the goal, which influences the trajectory only through the actions. By the chain rule,
\begin{equation}
    \mathbb{I}_{\pih,\pir}(\mathbf{A}_h \to \traj \mid s) \;=\; \Rev(s) \;+\; \mathbb{I}_{\pih,\pir}(\mathbf{A}_h \to \traj \mid G, s),
    \label{eq:massey-identity}
\end{equation}
whose second term is nonnegative. The undirected chain $G \to \mathbf{A}_h \to \traj$ is not Markov under feedback, because actions are common effects of the goal and past states, so the plain data-processing inequality is unavailable and is replaced by the directed form (Appendix~\ref{app:proofs} gives a two-line counterexample). The upper inequality holds because the human's realised behaviour, with the goal marginalised out, is itself a causally conditioned input distribution, so it competes in the maximisation defining $\empow(s)$.

\begin{lemma}[Monotone floor]
\label{lem:monotone-floor}
$\Bs$ is a strictly increasing function of $I_h(s)$ on $I_h(s) \ge 0$, and $\Bs \to \log_2 \Ngoals$ as $I_h(s) \to \log_2 \Ngoals$.
\end{lemma}

Maximising individual power therefore maximises a guaranteed floor on revelation. In the attainable-capability limit, the floor meets the ceiling, i.e., revelation is forced to its maximum wherever the environment permits perfect capability. This is the sense in which attainment implies revelation. The bound applies only in this direction (Section~\ref{sec:separations} shows how both converse directions fail). Capabilities are evaluated under whatever goal-conditioned policy $\pih$ is in play, optimal or not, so suboptimal pursuit and environmental noise lower $I_h(s)$ rather than break the bound.

Each rung of \eqref{eq:bridge-chain} answers a different question:
\begin{center}
\small
\begin{tabular}{lll}
\toprule
quantity & question it answers & goal variable \\
\midrule
$I_h(s)$, $\Bs$ & how attainable the candidate goals are & counterfactual \\
$\Rev(s)$ & how much outcomes identify the pursued goal & latent \\
$\mathbb{I}(\mathbf{A}_h \to \traj \mid s)$ & how much influence-information pursuit \emph{does} generate & marginalised \\
$\empow(s)$ & how much influence \emph{could} be exercised & none \\
\bottomrule
\end{tabular}
\end{center}
None of the above is ``the amount of control the human exerted''. High $\Rev(s)$ asserts that outcomes distinguish goals, not that they are achieved. And the rung that looks most like control, the directed information, has its test in the deterministic pursuer of Remark~\ref{rem:deterministic-pursuit}. Given their goal, nothing in their actions is left to chance, yet they generate no directed information at all. The quantity counts variation in the actions that reaches the trajectory, not the actions themselves.

\paragraph{Two gaps.} By \eqref{eq:massey-identity}, the distance from capacity to revelation splits as
\begin{equation}
    \empow(s) - \Rev(s) \;=\; \underbrace{\big(\empow(s) - \mathbb{I}(\mathbf{A}_h \to \traj \mid s)\big)}_{\text{capacity margin}} \;+\; \underbrace{\mathbb{I}(\mathbf{A}_h \to \traj \mid G, s)}_{\text{goal-blind influence}}.
    \label{eq:two-gaps}
\end{equation}
The first term is the capacity unachieved by the realised ensemble. The capacity-achieving input is a randomisation device, rather than a rational goal pursuit policy, and a goal-pursuer needs at most $\log_2 \Ngoals$ bits of the channel however large its capacity, meaning that the capacity margin is not a deficiency of the human's behaviour. The margin measures how far the environment's influence budget exceeds what goal-conditioned behaviour uses. Section~\ref{sec:separations} shows a maximiser of $\empow$ chases this margin. The second term is directed information the actions generate that the goal does not explain, i.e. randomness in pursuit itself such as due to exploration, noise, mixed strategies. It vanishes for deterministic pursuit.

\begin{remark}[Deterministic pursuit]
\label{rem:deterministic-pursuit}
If each $\pih(\cdot \mid \cdot, g)$ is deterministic, then given $G$ the actions are a function of the realised states, so $\mathbb{I}(\mathbf{A}_h \to \traj \mid G, s) = 0$ and
$\Rev(s) = \mathbb{I}(\mathbf{A}_h \to \traj \mid s)$. In this case, every bit of influence-information the interaction generates is a revealed-goal bit, and the whole distance to capacity is margin. Rational pursuit tends to deterministic policies, so this is the typical regime, and it is why maximising directed information and pursuing a goal are opposite activities: the first requires randomisation which the second avoids in the limit.
\end{remark}

\paragraph{Aliasing and ambiguity.} There are two reasons that can make revelation fall short of the ceiling: distinct goals may induce the same behaviour, and distinct behaviours may produce indistinguishable trajectories. Both are undistinguished by $\mathbb{H}(G \mid \traj, s)$, but they separate cleanly through the behaviour the goal induces.

\begin{definition}[Behaviour class]
\label{def:behaviour-class}
Fix $s$ and $(\pih, \pir)$, and let $\mathcal{H}_s$ be the set of histories with positive probability from $s$ under some $g \in \Gh$. Say $g \sim_s g'$ when $\pih(\cdot \mid \eta, g) = \pih(\cdot \mid \eta, g')$ for every $\eta \in \mathcal{H}_s$, and let $\Zpol \coloneqq [G]_{\sim_s}$ be the \emph{behaviour class}: the goal equivalence class that removes those differences in induced policy that are not shown by reachable histories.
\end{definition}

A goal reaches the trajectory only through the policy it induces, so the trajectory reveals no more and no less of the goal than of its class, and $\mathbb{I}(G;\traj \mid s)$ equals $\mathbb{I}(\Zpol;\traj \mid s)$. The class provides a split of the equivocation into its two causes.

\begin{proposition}[Aliasing--ambiguity decomposition]
\label{prop:z-decomposition}
Under Assumption~\ref{ass:goal-prior}, $G \to \Zpol \to \traj$ is a Markov chain, and
\begin{equation}
    \mathbb{H}(G \mid \traj, s) = \underbrace{\mathbb{H}(G \mid \Zpol, s)}_{\textnormal{goal--policy aliasing}} + \underbrace{\mathbb{H}(\Zpol \mid \traj, s)}_{\textnormal{policy--trajectory ambiguity}},
    \qquad
    \Rev(s) = \mathbb{I}(\Zpol; \traj \mid s) = \mathbb{H}(\Zpol \mid s) - \mathbb{H}(\Zpol \mid \traj, s).
    \label{eq:z-decomposition}
\end{equation}
In particular $\Rev(s) \le \mathbb{H}(\Zpol \mid s)$: revelation is capped by the entropy of the behaviour ensemble, not of the goal ensemble.
\end{proposition}

If a hundred goals induce two observable behaviours,interaction can reveal at most one bit, however deterministic and however well executed. 
In Equation~\eqref{eq:two-gaps}, aliasing is a refinement of the equivocation.
Under deterministic pursuit, where the goal-blind term vanishes, its effect appears entirely as capacity margin, since fewer distinct behaviours generate less influence-information. 
Both terms of \eqref{eq:z-decomposition} are computed exactly in the experiment of Section~\ref{sec:experiment}.

\section{Both Converses Fail}
\label{sec:separations}

In the chain~\eqref{eq:bridge-chain}, each quantity can be large while the one below it stays at zero. This section makes the two failure modes behind that concrete: capacity without revelation, and revelation without attainment.

\subsection{Capacity Without Revelation}
\label{sec:capacity-separation}

Human goals are few and coarse relative to the environments that realise them. For instance, we care whether the room is clean, not about the position of each dust particle. This means that $\log_2 \Ngoals \ll \log_2 |\mathcal{S}|$ in any rich environment, while the entropy of the action-sequence variable, and with it the bound~\eqref{eq:empowerment} on $\empow$, grows linearly in the horizon. A capacity maximiser is therefore rewarded for influence that is not used by any  candidate goal. The separation below realises this as an argmax reversal between two robot policies.

\begin{proposition}[Capacity without revelation]
\label{prop:capacity-separation}
For every $\Ngoals \ge 2$, $m \ge 1$, and horizon $T \ge 1$ there is a fully observable, deterministic environment with initial state $s$, a goal family of $\Ngoals$ first-outcome goals satisfying Assumptions~\ref{ass:goal-prior}--\ref{ass:atomic-goals} with $\gamma_h = 1$, and two robot policies\footnote{Formally the robot has two actions, \textsf{gate-goals} and \textsf{gate-toggles}, and the kernel encodes the contingency: under \textsf{gate-goals} every $\textsf{go}_j$ moves and every $\textsf{flip}_i$ is inert, and conversely. The two policies are the two constant deterministic ones.} (the prior uniform), $\pir^{\mathrm{tog}}$ (a board of $m$ free binary toggles responds to the human; goal transitions are blocked) and $\pir^{\mathrm{ass}}$ (goal transitions respond; toggles are frozen), such that when the human pursues their goal:
\begin{enumerate}[itemsep=0pt]
    \item $\empow(s \mid \pir^{\mathrm{tog}}) = T\log_2(m+1)$ while $\Rev(s \mid \pir^{\mathrm{tog}}) = 0$ and every capability is $0$;
    \item $\empow(s \mid \pir^{\mathrm{ass}}) = \log_2(\Ngoals T + 1)$ while $\Rev(s \mid \pir^{\mathrm{ass}}) = \log_2 \Ngoals$, its ceiling, and every capability is $1$.
\end{enumerate}
Whenever $m \ge \Ngoals + 1$, $T\log_2(m+1) > \log_2(\Ngoals T + 1)$, so the capacity score prefers $\pir^{\mathrm{tog}}$, by a margin unbounded in $m$ and $T$, while its preferred policy leaves revelation, and every capability, at zero.
\end{proposition}

\begin{table}[htbp]
\centering
\footnotesize
\begin{tabular}{l c c}
\toprule
robot policy at $s$ & $\empow(s)$ & $\Rev(s)$ \\
\midrule
$\pir^{\mathrm{tog}}$ (toggle board) & $3\log_2 8 = 9$ bits & $0$ \\
$\pir^{\mathrm{ass}}$ (assistance)  & $\log_2 13 \approx 3.70$ bits & $\log_2 4 = 2$ bits \\
\bottomrule
\end{tabular}
\caption{Proposition~\ref{prop:capacity-separation} at $\Ngoals = 4$ goals, $m = 7$ toggles, $T = 3$ steps: capacity prefers the board by $\approx 5.30$ bits; under that policy the trajectory carries nothing about the goal.}
\label{tab:capacity-instance}
\end{table}

Table~\ref{tab:capacity-instance} shows an instance of this proposition. The board capacity, deterministic and reversible, is worthless to every goal in $\Gh$, but the capacity score does not account for it. This does not depend on modelling the human as a pure goal-pursuer, since a human who repurposed the board to signal their goal could push $\Rev$ no higher than its ceiling $\log_2 \Ngoals = 2$ bits, which the board capacity dwarfs. On this instance a deterministic pursuer generates no goal-blind influence, and under $\pir^{\mathrm{tog}}$ generates no influence-information at all, so the gap decomposition~\eqref{eq:two-gaps} is all margin.

\subsection{Revelation Without Attainment}
\label{sec:scramble-separation}

Under the uniform prior, revelation is a functional of the family of goal-conditioned trajectory laws $\{p(\cdot \mid G = g, s)\}_{g \in \Gh}$, invariant under relabelling which law goes with which goal. Attainment, in contrast, is a property of the goal--law pairing, asking whether the law paired with $g$ reaches $g$'s satisfaction set.

\begin{proposition}[Revelation without attainment]
\label{prop:scramble-separation}
Fix $s$, policies $(\pih, \pir)$ inducing laws $p(\traj \mid G = g, s)$ and capabilities $\Ch(s, g \mid \pig)$. For a permutation $\sigma$ of $\Gh$, call a configuration $\sigma$-scrambled if it induces the laws $p_\sigma(\traj \mid G = g, s) = p(\traj \mid G = \sigma(g), s)$ for every $g$: the human pursues $g$ through the same policy, and the interface delivers what pursuing $\sigma(g)$ would have delivered. Under Assumptions~\ref{ass:goal-prior}--\ref{ass:atomic-goals} with the uniform prior:
\begin{enumerate}[itemsep=0pt]
    \item $\Rev$ is invariant: $\mathbb{I}_\sigma(G; \traj \mid s) = \mathbb{I}(G; \traj \mid s)$;
    \item if $p(\traj \in \Trajs_{g'} \mid G = g, s) = 0$ for every $g' \neq g$ (i.e., under each original law, pursuit never attains a goal other than the one pursued), then the capability of $g$ under the scrambled law, $\mathbb{E}_{p_\sigma(\cdot \mid G = g, s)}\big[\gamma_h^{\tau_g(\traj)}\,\mathbf{1}(\traj \in \Trajs_g)\big]$, equals $\Ch(s, g \mid \pig)$ when $\sigma(g) = g$ and is $0$ otherwise. A fixed-point-free $\sigma$ zeroes every capability, sending $I_h(s) \to -\infty$ and $\Bs \to 0$.
\end{enumerate}
Consequently, the floor of Theorem~\ref{thm:bridge} has no converse: no function of $\Rev(s)$, or of any statistic invariant under goal relabelling, bounds capability from below. Configurations exist with identical revelation and capability profiles $(c, \dots, c)$ versus $(0, \dots, 0)$.
\end{proposition}

\begin{table}[htbp]
\centering
\footnotesize
\begin{tabular}{l c c c}
\toprule
interface at $s$ & $\Rev(s)$ & capability profile & floor $\Bs$ \\
\midrule
aligned & $0.9$ bits & $(0.9,\; 0.9)$ & $0.531$ bits \\
$\sigma$-scrambled (swap) & $0.9$ bits & $(0,\; 0)$ & $0$ \\
\bottomrule
\end{tabular}
\caption{Proposition~\ref{prop:scramble-separation} at $\Ngoals = 2$ with $\sigma$ the transposition; the aligned interface realises each intended goal with probability $c = 0.9$ and otherwise ends in a goal-independent failure trajectory. Whilst revelation is the same in both cases, the human attains their goal $90\%$ of the time in one and never in the other.}
\label{tab:scramble-instance}
\end{table}

A minimal case in Table~\ref{tab:scramble-instance} illustrates how $\mathbb{I}(G;\traj \mid s)$ measures identifiability rather than control. Additionally, the satisfaction decoder behind Theorem~\ref{thm:bridge} checks $\traj \in \Trajs_g$ for the goal actually pursued, so a scramble sends its correct-guess probability (and with it the floor) to zero, while $\Rev$ stays unchanged. The guarantee runs from attainment upward.

\section{A Floor for Overlapping Goal Families}
\label{sec:overlap}

The Fano floor \eqref{eq:fano-floor} requires disjoint satisfaction sets (Assumption~\ref{ass:atomic-goals}) and parameters that are counterfactual pursuit probabilities. We now derive a second floor without these requirements. It reads the trajectory through one bit (whether it satisfies the goal pursued), and bounds $\Rev(s)$ by what that bit alone can certify. The restriction discards timing and route information by design, which makes the bound depend on the interaction only through two scalar statistics: the \emph{efficacy} $\Pbar(s)$, the probability that the pursued goal is satisfied within the window, and the \emph{specificity} $\bar{K}(s)$, the expected number of goals a trajectory satisfies, intended or not. The ratio $\bar K(s)/\Ngoals$ is the probability that a trajectory satisfies a goal by chance, rather than under pursuit.

\begin{theorem}[Satisfaction floor]
\label{thm:vim}
Under the uniform prior assumption (\ref{ass:goal-prior}), for any finite family of trajectory goals, possibly overlapping,
\begin{equation}
    \Rev(s) \;\ge\; d\!\left(\Pbar(s) \,\middle\|\, \frac{\bar K(s)}{\Ngoals}\right),
    \qquad d(p \,\|\, q) \coloneqq p \log_2 \frac{p}{q} + (1-p)\log_2\frac{1-p}{1-q},
    \label{eq:vim-bound}
\end{equation}
the binary relative entropy between the hit rate under pursuit and the hit rate under chance, with the convention $0 \log_2 0 = 0$.
\end{theorem}

Expanding $d$ and collecting the $\log_2 \Ngoals$ terms,
\begin{equation}
    \Rev(s) \;\ge\; \log_2 \Ngoals - h_2(\Pbar(s)) - \Pbar(s)\log_2 \bar{K}(s) - \big(1 - \Pbar(s)\big)\log_2\big(\Ngoals - \bar{K}(s)\big),
    \label{eq:vim-expanded}
\end{equation}
with $h_2$ the binary entropy. Read left to right: the ceiling, less the entropy of the success bit, less the expected log-count of goals consistent with that bit, $\bar K$ of them on success and $\Ngoals - \bar K$ on failure.

We now consider the limit cases for a sanity check. Satisfying only the pursued goal ($\Pbar = 1$, $\bar K = 1$) gives $d(1 \,\|\, 1/\Ngoals) = \log_2 \Ngoals$, the ceiling. Behaviour independent of the goal, satisfying $\bar K$ goals by accident ($\Pbar = \bar K / \Ngoals$), gives $0$. Avoiding the pursued goal while satisfying $\bar K$ others ($\Pbar = 0$) gives $\log_2\!\big(\Ngoals / (\Ngoals - \bar K)\big) > 0$, which again shows that revelation is not a desired target in itself. The sign of $\Pbar - \bar K/\Ngoals$ records which way pursuit departs from chance.

\begin{remark}[The same floor as a variational bound]
\label{rem:vim-decoder}
For any decoder $q(g \mid \traj)$, the variational lower bound of \citeauthor{Barber2003Variational} \cite{Barber2003Variational}, whose use for intrinsic control is due to \citeauthor{Mohamed2015VIM} \cite{Mohamed2015VIM}, gives
\begin{equation}
    \Rev(s) \;\ge\; \mathbb{H}(G \mid s) + \mathbb{E}_{G, \traj}\big[\log_2 q(G \mid \traj)\big].
    \label{eq:vim-generic}
\end{equation}
A softmax over the satisfaction indicators $\mathbf{1}(\traj \in \Trajs_g)$ with one log-odds weight, optimised over that weight, returns Equation~\eqref{eq:vim-bound} (Appendix~\ref{app:proofs}). The hard decoder of Theorem~\ref{thm:bridge} is its infinite-weight limit for a disjoint goal family. Theorem~\ref{thm:vim} is therefore also the best bound that decoder class can give, where the sign of its optimal weight is the sign of $\Pbar - \bar K/\Ngoals$.
\end{remark}

The two floors connect through the capabilities. The average capability $\Cbar(s) = \frac{1}{\Ngoals}\sum_g \Ch(s, g \mid \pig)$ lower-bounds the efficacy $\Pbar(s)$, since discounting and the first-outcome charge only subtract, and substituting $\Cbar$ for $\Pbar$ in \eqref{eq:vim-bound} is conservative (when $\Cbar(s) \ge \bar K(s)/\Ngoals$, the region where $d$ increases in its first argument). Unlike the counterfactual capabilities, $(\Pbar, \bar K)$ are population statistics of observable interactions: given per-goal satisfaction labels (which for goals expressed as regular properties a monitor automaton computes directly from logged trajectories \cite{ToroIcarte2022RewardMachines}) and the pursued goal per window for $\Pbar$, both are means of bounded per-window statistics, so finite-log estimates concentrate at standard rates. Equation~\eqref{eq:vim-bound} then gives a computable lower bound on the goal revelation a deployed system's users retain, providing a quantitative handle on the gradual disempowerment concern \cite{Kulveit2025Gradual}. However, the estimate is only as good as the labels, is relative to the family $\Gh$ and the window segmentation, and is a measurement, not a training signal.

\section{Numerical Illustration}
\label{sec:experiment}

Section~\ref{sec:separations} included simple designed environments to illustrate failure modes. This section asks what the chain looks like in generic environments: how much of capacity becomes revelation, how much revelation the floor certifies, and how far the three scores agree about which state is best (with no adversarial aspects).

We focus on a multi-armed bandit as the simplest setting exhibiting the properties above. The goals are the outcomes themselves, a disjoint family by construction. At horizon one a trajectory is a single transition, so everything the three measures see of a state is its induced channel~\eqref{eq:induced-channel}. That channel is an $m \times \Ngoals$ stochastic matrix from the human's $m$ actions to $\Ngoals$ outcome states. We therefore identify a state with that matrix, writing $P(g \mid a)$ for the probability that action $a$ leads to outcome $g$, and sample states by drawing each row from a Dirichlet$(\alpha)$ distribution. Pursuit is rational and deterministic, $a^{*}(g) \in \argmax_a P(g \mid a)$ (ties to the first index), so the capability is $\Ch(g) = \max_a P(g \mid a)$. The individual power $I_h$ is taken at $\zeta = 1$, $\gamma_h = 1$. Capacity $\empow$ is computed exactly as the channel capacity by Blahut--Arimoto with a duality-gap certificate of $10^{-8}$ nats, cross-checked against an independent solver to $1.1 \times 10^{-10}$ bits on the first 150 channels. The behaviour-class terms from the joint laws, $\Zpol$, is the chosen arm $a^{*}(G)$. The canonical configuration is $\Ngoals = 5$, $m = 8$, $\alpha = 1$, 1{,}000 independent environments of 50 states each, from a fixed seed\footnote{Code and seed accompany the paper (\texttt{experiment/bridge\_experiment.py}, seed 20260823; see Code availability), and the code regenerates the per-state outputs.}.
The simulations highlight three regularities and one caution:
\begin{itemize}
    \item The ordering $\Bs \le \Rev(s) \le \mathbb{I}(\mathbf{A}_h; S_1 \mid s) \le \empow(s)$ holds on every one of the 50{,}000 states. With deterministic pursuit, every bit of realised influence-information is a revealed-goal bit (to $6 \times 10^{-16}$, illustrating Remark~\ref{rem:deterministic-pursuit}). On average the states carry $\empow(s) = 0.585$ bits of capacity, of which $\Rev(s) = 0.497$ bits (85.0\%) is revelation, of which $\Bs = 0.274$ bits (54.6\%) is certified by the floor from $I_h = 1.228$.
    
    \item Illustrating the equivocation split of Proposition~\ref{prop:z-decomposition}, out of $\log_2 5 \approx 2.32$ bits of prior entropy, ambiguity accounts for $1.687$ bits on average and aliasing for $0.138$. Aliasing is positive in 33.1\% of states (two goals sharing an argmax arm) and quantised, equal to $2/5$ of a bit whenever one pair collides at $\Ngoals = 5$. In this ensemble the ceiling $\mathbb{H}(\Zpol \mid s)$ binds a third of the time, with most of what trajectories fail to reveal being channel noise rather than behavioural coincidence.

\begin{table}[htbp]
\centering
\footnotesize
\begin{tabular}{lccc}
\toprule
 & $(\empow, \Rev)$ & $(\Rev, B)$ & $(\empow, B)$ \\
\midrule
Mean Kendall $\tau$ (within an env't) & $0.751$ & $0.619$ & $0.581$ \\
Exact argmax agreement & $57.8\%$ & $41.8\%$ & \\
\bottomrule
\end{tabular}
\quad
\begin{tabular}{ccccc}
\toprule
$\Ngoals$ & $m$ & $\alpha$ & $\tau(\empow, \Rev)$ & $\tau(\Rev, B)$ \\
\midrule
3 & 4 & 1.0 & $0.863$ & $0.693$ \\
5 & 8 & 1.0 & $0.754$ & $0.617$ \\
8 & 12 & 1.0 & $0.700$ & $0.556$ \\
5 & 8 & 0.2 & $0.809$ & $0.715$ \\
5 & 8 & 5.0 & $0.746$ & $0.569$ \\
\bottomrule
\end{tabular}
\caption{Left: rank agreement across the 50 states of an environment, averaged over 1{,}000 environments (canonical configuration). Right: sensitivity of the mean $\tau$ to the configuration, 300 environments each. Agreement falls as the spaces grow and rises for spikier channels ($\alpha = 0.2$).}
\label{tab:experiment}
\end{table}

    \item Rank agreement (Table~\ref{tab:experiment}): the three scores ($\empow$, $\Rev$, $B$) co-vary, with mean within-environment Kendall $\tau$ between $0.58$ and $0.75$, yet exact argmax agreement is close to a coin flip, $57.8\%$ between $\empow$ and $\Rev$ and $41.8\%$ between $\Rev$ and $B$. Near-argmax behaviour is tighter, as the state maximising $B$ sits on average at the 96.4th percentile of $\Rev$ (top decile in $88.2\%$ of environments), and the state maximising $\Rev$ at the 98.3rd percentile of $\empow$. Maximising the floor is a good, not perfect, proxy for maximising revelation in generic environments. The sensitivity sweep shows every agreement falling as the outcome and action spaces grow.

    \item The caution is that this ensemble is favourable to agreement by construction. Every outcome is a candidate goal, so there are no degrees of freedom worthless to the family. The goal--outcome correspondence is identity, so nothing is scrambled. These are the two mechanisms of Section~\ref{sec:separations}, and introducing either one can drive the corresponding agreement to zero. The simulation adds the complement of the separations: even with both mechanisms absent, the quantities are correlated but far from interchangeable, and the scores do not agree on which state is best.
\end{itemize}

\section{Related Work}
\label{sec:related}

Mutual-information objectives for agents sort by three choices: which pair of variables the information couples, who holds the latent, and whether the quantity is maximised or minimised. The chain of Section~\ref{sec:bridge} falls under the ``goal to trajectory'', ``exogenous human goal'' latent, and an assistant (rather than the goal's owner) maximising the channel. The lines below differ from our bounds in at least one of the three choices.

\paragraph{Empowerment.} Klyubin et al.\ \cite{Klyubin2005EmpowermentUniversal} introduced Empowerment as $n$-step open-loop channel capacity from actions to a future state, as task-agnostic intrinsic motivation; Salge et al.\ \cite{Salge2014Empowerment} survey the formal properties. Our closed-loop form follows the directed-information treatment of causal information flow in sequential settings \cite{massey1990causality, Tiomkin2017Unified}, and the feedback-aware process Empowerment of \citeauthor{Tiomkin2025ProcessEmpowerment} \cite{Tiomkin2025ProcessEmpowerment}. Proposition~\ref{prop:capacity-separation} turns the informal worry that capacity is goal-blind into an argmax reversal against a revelation-maximal alternative.

\paragraph{Empowerment as assistance.} Pointing the quantity at the human predates the algorithms: Salge and Polani \cite{Salge2017ThreeLaws} propose a language-free stand-in for Asimov's laws that combines three Empowerment heuristics, one of which keeps the human's Empowerment high. Du et al.\ \cite{Du2020AvE} propose human Empowerment as an auxiliary reward term for an assistant, and implement not the channel capacity but a proxy based on the diversity of reachable final states. Franzmeyer et al.\ \cite{Franzmeyer2022Altruistic} cast altruism as preserving another agent's choice, with reachable-state diversity again standing in. Myers et al.\ \cite{Myers2024Assist} scale the commitment with contrastive successor representations, maximising the influence of the human's actions on future states without inferring rewards. This line of work keeps the human side goal free, so Proposition~\ref{prop:capacity-separation} applies where controllable degrees of freedom outnumber candidate goals, and Proposition~\ref{prop:scramble-separation} applies to any surrogate invariant under relabelling of the goal--outcome correspondence.

\paragraph{Skill discovery.} In this family of prior art, the latent is owned by the agent itself. Variational Intrinsic Control \cite{Gregor2016VIC} maximises $\mathbb{I}(\Omega; S_f \mid s_0)$ over options $\Omega$ with the option prior optimised, read by its authors as Empowerment over a learned option channel. Skew-Fit skews a goal prior toward rarely visited states \cite{Pong2020SkewFit}. DADS \cite{Sharma2020DADS} maximises a one-step $\mathbb{I}(Z; S' \mid s)$ with the prior fixed, DIAYN \cite{Eysenbach2019DIAYN} a per-state discriminator variant, and Choi et al.\ \cite{Choi2021VariationalEmpowerment} subsume goal-conditioned reinforcement learning under variational Empowerment. Eysenbach et al.\ \cite{Eysenbach2022Geometry} prove that mutual-information skill discovery does not learn skills optimal for every possible reward function, which is congruent with Section~\ref{sec:separations}. They also show that, under an assumed adaptation procedure, the learned skill distribution is the initialisation that minimises regret against adversarially chosen rewards. METRA replaces the MI objective outright after diagnosing that it misaligns with useful behavioural spread \cite{Park2024METRA}. For a self-assigned latent, relabelling is a symmetry and is not an issue. When applied to another agent's goal variable, whose labels carry commitments the assistant is meant to serve, the same invariance is Proposition~\ref{prop:scramble-separation}. The variational decoders of that literature are trained to tighten the bound, whereas the decoder of Section~\ref{sec:overlap} is restricted to satisfaction indicators so that what it certifies is estimable.

\paragraph{Intent legibility.} On the actor's side, legible motion \cite{Dragan2013Legibility} maximises an observer's posterior on the actor's \emph{actual} goal from a trajectory prefix. Strouse et al.\ \cite{Strouse2018Intentions} add or subtract the goal information in actions, $\mathbb{I}(A; G \mid S)$, or in states, $\mathbb{I}(S; G)$, to share or hide intent. 
The same information is treated as a cost to be minimised in
van Dijk and Polani's relevant goal information (the minimal goal information a policy must consult, whose spikes mark natural subgoals) \cite{VanDijk2011Subgoals}, InfoBot's $\mathbb{I}(A; G \mid S)$ bottleneck for transferable defaults \cite{Goyal2019InfoBot}, and opacity enforcement (minimising what observations leak about a secret) \cite{Shi2024Opacity}. These works either anchor to the true goal, which immunises them against relabelling, or treat the actor's own latent, for which relabelling is again a symmetry. The satisfaction decoder behind Theorem~\ref{thm:bridge} is anchored in Dragan's sense, checking the trajectory against the goal actually pursued. A relabelling therefore leaves the aggregate mutual information unchanged while sending the decoder's success rate, and with it the floor, to zero.

\paragraph{Interface optimisation.} Reddy et al.\ \cite{Reddy2022FirstContact} point out that randomly permuting the keybindings of an assistive interface leaves the one-step information $\mathbb{I}(a; s' \mid s)$ from the user's commands to the next state unchanged. This is a one-step instance of the invariance in Proposition~\ref{prop:scramble-separation}. They offer a behavioural solution, adding $\mathbb{I}(s; a)$ to the objective on the assumption that co-adapting users issue noisier commands through unintuitive interfaces. Capability anchoring is a structural alternative that collapses under permutations with no assumption about user noise.

\paragraph{Multi-agent information objectives.} Multi-agent reinforcement learning has a dense family of information objectives, with latents held by the team. MAVEN couples one shared latent to the joint trajectory for committed exploration \cite{Mahajan2019MAVEN}, and role and diversity methods couple agent-specific latents to that agent's own trajectory or observations \cite{Wang2020ROMA, Li2021CDS, Jiang2021EOI}, extended hierarchically \cite{Yang2023HMASD}. From offline data, Chen et al.\ \cite{Chen2025VOMASD} learn hierarchical skill codes with an auto-encoder rather than an information objective. Social influence puts the information between actions instead, rewarding $\mathbb{I}(a_i; a_j)$ \cite{Jaques2019Influence}, routed through a shared coordination latent by Kim et al.\ \cite{Kim2023VM3AC}. On the perception side, IMAS$^2$ couples a selected subset of agents' observations to the joint trajectory \cite{Shi2026IMAS}. Multi-human assistance exists in goal-free or information-free forms: coupled Empowerment maximisation combines the companion's and the player's capacities with a transfer capacity between them \cite{Guckelsberger2016Coupled}, multi-principal assistance games aggregate utilities by social choice with no information quantity \cite{Fickinger2020MPAG}, and a recent preprint reports that an assistant maximising one human's Empowerment measurably reduces a bystander's in 27--56\% of procedurally generated environments \cite{Yang2025Bystander}. All of these are per-agent or goal-free constructions, so neither mechanism of Section~\ref{sec:separations} is expressible in them, and a tuple of distinct private goals read through one shared trajectory has no occupant. The single-human chain proved here is the base case for that extension.

\paragraph{Capability-based power.} Heitzig and Potham \cite{Heitzig2026FairObjective} construct individual power axiomatically from goal-attainment capabilities and argue for soft maximisation of its long-term aggregate as the robot's objective, goal-agnostic by design. We adopt their capability as the attainment rung of the chain and prove, in information terms, that it provides a floor under revelation. Turner et al.\ \cite{Turner2021Power} prove optimal policies tend to seek power for the \emph{agent}. In contrast, the chain \eqref{eq:bridge-chain} is a formal account of the complementary quantity, the power the \emph{human} retains, and Kulveit et al.\ \cite{Kulveit2025Gradual} motivate measuring its erosion, for which the estimable floor of Section~\ref{sec:overlap} is one candidate instrument.

\section{Discussion}
\label{sec:discussion}

Theorem~\ref{thm:bridge} orders individual power, goal revelation and empowerment in a chain in which attainment implies revelation, and Section~\ref{sec:separations} shows that both converses fail.
The bridge licenses a floor statement, meaning that maximising individual power maximises a guaranteed lower bound on goal revelation (Lemma~\ref{lem:monotone-floor}), a bound that meets the ceiling where the environment permits perfect capability.
It does not license ``maximising $I_h$ maximises $\mathbb{I}(G;\traj \mid s)$'', which is false in general.
The two floors admit a division of labour.
The Fano floor is a design-time statement in the objective's own variable, with counterfactual parameters that a log cannot estimate, whilst the satisfaction floor is a deployment-time measurement in statistics that logs do estimate.
Rewarding a system on its own measurement corrupts the measurement, so the satisfaction floor is a monitor rather than a training signal.

\paragraph{Limitations.} All quantities live on a finite window: goals are trajectory properties certified within $T$ steps, which covers reach, pattern, and within-window maintenance goals but not unbounded safety properties, whose attainment lies outside the finitely-certifiable class \cite{Yang2022LTLIntractable}. Full observability is assumed throughout: human, robot, and the decoding observer all see the state, and partial observability changes both the capability recursion and the decoding argument. The Fano route needs disjoint satisfaction sets, manufactured by the first-outcome construction, with the satisfaction floor as the instrument for genuinely overlapping families. Every information quantity is relative to the reference prior and the candidate family: a different $\Gh$ gives different numbers, and we did not consider goals outside the family. Capability is counterfactual, conditioning the human's policy on goals they may not hold, so a robot that computes $I_h(s)$ must model that policy. Theorem~\ref{thm:bridge} then bounds the revelation of the modelled human, and carries over to the actual human only as far as the model is accurate. The human holds one goal per window, drawn afresh across windows. Graded mixtures of goals within a window are outside the present decoding argument, and the soft decoder of Section~\ref{sec:overlap} is the natural starting point for that extension.

\section{Conclusion}
\label{sec:conclusion}

We set out to relate three measures of a human's power at a state, i.e. empowerment, goal revelation, and individual power. Theorem~\ref{thm:bridge} orders them in a chain. Fano's inequality turns individual power into a floor under revelation, so a robot that maximises individual power maximises a guaranteed lower bound on how much the interaction reveals about the goal pursued. Neither of the other two quantities provides such a guarantee. The capacity score includes influence over degrees of freedom that are not used by any candidate goal, and can prefer an interface on which every candidate goal is unattainable (Proposition~\ref{prop:capacity-separation}). Revelation measures whether trajectories discern goals rather than whether goals are attained, so a relabelling that zeroes every capability leaves it unchanged (Proposition~\ref{prop:scramble-separation}). This means that positive guarantees about the power a human retains come only from the attainment end of the chain. We make no claim that any of the three measures is a safe objective on its own.

The results provide four things beyond the chain itself.
\begin{itemize}
    \item The assumption that maximising empowerment preserves the human's ability to attain their goals does not hold in general. Proposition~\ref{prop:capacity-separation} gives environments in which the empowerment-maximising interface leaves every candidate goal unattainable, by a margin that grows without bound in the number of goal-irrelevant degrees of freedom.

    \item Proposition~\ref{prop:z-decomposition} separates the two reasons a trajectory fails to reveal intent, i.e. goal--policy aliasing, when distinct goals induce the same behaviour, and policy--trajectory ambiguity, when distinct behaviours produce indistinguishable trajectories. This shows that revelation is capped by the entropy of the behaviour ensemble rather than that of the goal ensemble.

    \item The satisfaction floor (Theorem~\ref{thm:vim}) requires two bounded population statistics, the efficacy $\Pbar(s)$ and the specificity $\bar K(s)$, both of which are means of per-window quantities that interaction logs record. This provides a computable lower bound on the goal revelation that the users of a deployed system retain, i.e. one quantitative handle on gradual disempowerment, which does not require the counterfactual capabilities that a log cannot estimate.

    \item Skill-discovery objectives optimise the same mutual information with the latent held by the agent itself, for which relabelling is a symmetry. Pointed at a human's goals, the same objective inherits the invariance of Proposition~\ref{prop:scramble-separation}, so an assistant optimising it cannot distinguish a working interface from a rewired one.
\end{itemize}

Three directions follow from the limitations in Section~\ref{sec:discussion}. The capability objective inherits the choice of its goal family, and a robot maximising it can game that choice. We study this failure and an architectural repair in a companion manuscript.{} The results assume full observability and a finite interaction window, and relaxing either changes both the capability recursion and the decoding argument. Finally, the floors are stated for the robot's model of the human's policy, so a model refined from interaction, with a bound on its error, would carry the guarantees over to the actual human.

\section*{Code availability}
The code for Section~\ref{sec:experiment} and a script that recomputes the numbers quoted in this paper are available at \url{https://github.com/RuizSerra/revealing-goals}.

\section*{Use of AI tools}
We used large language models throughout this work: Claude (Fable 5, Fable 5.1, Opus 5 and Opus 5.5; Anthropic) and Gemini Pro 3.8 (Google).
The work began with an extended, interactive exchange between one of the authors and Claude, in which the research questions, framing and candidate results were developed jointly over many iterations.
In this phase, before any drafting, the author set the direction of the work, deciding which lines of inquiry to pursue and which results to retain, refine or set aside.
Claude subsequently drafted the text, carried out the mathematical derivations, and wrote the code and tables, while the author continued to steer the content through successive revisions.
The same author then revised the prose of the whole main text for readability.
Fresh Claude instances and Gemini were used to red-team the mathematics, prose and claims, and to check that the cited works support the statements attributed to them.
The authors take full responsibility for the paper.

\section*{Acknowledgements}
This work was carried out as part of the PIBBSS Fellowship, with support from Principles of Intelligence, a not-for-profit organisation.

\section*{Author contributions}
Contributions follow the CRediT taxonomy.
\begin{description}[font=\normalfont, leftmargin=0pt, itemsep=2pt]
  \item[JRS:] Conceptualisation (lead); formal analysis (lead); funding acquisition; investigation (lead); project administration; software (lead); validation; visualisation; writing~-- original draft (lead); writing~-- review and editing.
  \item[TG:] Conceptualisation; formal analysis; supervision; validation; writing~-- review and editing.
  \item[JH:] Conceptualisation; formal analysis; investigation; software; supervision; validation; writing~-- review and editing.
\end{description}

\printbibliography

@inproceedings{Barber2003Variational,
  title={Information Maximization in Noisy Channels: A Variational Approach},
  author={Barber, David and Agakov, Felix V.},
  booktitle={Advances in Neural Information Processing Systems},
  volume={16},
  year={2003}
}

@inproceedings{Choi2021VariationalEmpowerment,
  title={Variational Empowerment as Representation Learning for Goal-Conditioned Reinforcement Learning},
  author={Choi, Jongwook and Sharma, Archit and Lee, Honglak and Levine, Sergey and Gu, Shixiang Shane},
  booktitle={International Conference on Machine Learning},
  series={Proceedings of Machine Learning Research},
  volume={139},
  pages={1953--1963},
  year={2021}
}

@inproceedings{Chen2025VOMASD,
  author    = {Chen, Jiayu and Lan, Tian and Aggarwal, Vaneet},
  title     = {Variational Offline Multi-Agent Skill Discovery},
  booktitle = {Proceedings of the Thirty-Fourth International Joint Conference on Artificial Intelligence (IJCAI)},
  pages     = {4833--4841},
  year      = {2025},
  doi       = {10.24963/ijcai.2025/538}
}

@inproceedings{Dragan2013Legibility,
  title={Legibility and Predictability of Robot Motion},
  author={Dragan, Anca D. and Lee, Kenton C. T. and Srinivasa, Siddhartha S.},
  booktitle={ACM/IEEE International Conference on Human-Robot Interaction},
  year={2013}
}

@inproceedings{VanDijk2011Subgoals,
  author    = {van Dijk, Sander G. and Polani, Daniel},
  title     = {Grounding Subgoals in Information Transitions},
  booktitle = {IEEE Symposium on Adaptive Dynamic Programming and Reinforcement Learning (ADPRL)},
  year      = {2011},
  doi       = {10.1109/ADPRL.2011.5967384}
}

@inproceedings{Du2020AvE,
  title={AvE: Assistance via Empowerment},
  author={Du, Yuqing and Tiomkin, Stas and Kiciman, Emre and Polani, Daniel and Abbeel, Pieter and Dragan, Anca},
  booktitle={Advances in Neural Information Processing Systems},
  volume={33},
  year={2020},
  pages={4560--4571}
}

@inproceedings{Eysenbach2022Geometry,
  author    = {Eysenbach, Benjamin and Salakhutdinov, Ruslan and Levine, Sergey},
  title     = {The Information Geometry of Unsupervised Reinforcement Learning},
  booktitle = {International Conference on Learning Representations},
  year      = {2022}
}

@inproceedings{Eysenbach2019DIAYN,
  title={Diversity is All You Need: Learning Skills without a Reward Function},
  author={Eysenbach, Benjamin and Gupta, Abhishek and Ibarz, Julian and Levine, Sergey},
  booktitle={International Conference on Learning Representations},
  year={2019}
}

@online{Fickinger2020MPAG,
  author     = {Fickinger, Arnaud and Zhuang, Simon and Critch, Andrew and Hadfield-Menell, Dylan and Russell, Stuart},
  title      = {Multi-Principal Assistance Games: Definition and Collegial Mechanisms},
  year       = {2020},
  eprint     = {2012.14536},
  eprinttype = {arxiv},
  doi        = {10.48550/arXiv.2012.14536},
  note       = {NeurIPS 2020 Cooperative AI Workshop; non-archival}
}

@inproceedings{Franzmeyer2022Altruistic,
  author    = {Franzmeyer, Tim and Malinowski, Mateusz and Henriques, Jo{\~a}o F.},
  title     = {Learning Altruistic Behaviours in Reinforcement Learning without External Rewards},
  booktitle = {International Conference on Learning Representations},
  year      = {2022}
}

@inproceedings{Goyal2019InfoBot,
  title={InfoBot: Transfer and Exploration via the Information Bottleneck},
  author={Goyal, Anirudh and Islam, Riashat and Strouse, Daniel J. and Ahmed, Zafarali and Larochelle, Hugo and Botvinick, Matthew M. and Bengio, Yoshua and Levine, Sergey},
  booktitle={International Conference on Learning Representations},
  year={2019}
}

@online{Gregor2016VIC,
  title      = {Variational Intrinsic Control},
  author     = {Gregor, Karol and Rezende, Danilo Jimenez and Wierstra, Daan},
  date       = {2016},
  eprint     = {1611.07507},
  eprinttype = {arXiv},
  doi        = {10.48550/arXiv.1611.07507},
  note       = {Presented in the ICLR 2017 Workshop Track}
}

@inproceedings{Guckelsberger2016Coupled,
  author    = {Guckelsberger, Christian and Salge, Christoph and Colton, Simon},
  title     = {Intrinsically Motivated General Companion {NPCs} via Coupled Empowerment Maximisation},
  booktitle = {IEEE Conference on Computational Intelligence and Games (CIG)},
  year      = {2016},
  doi       = {10.1109/CIG.2016.7860406}
}

@online{Heitzig2026FairObjective,
  ids         = {Heitzig2025ModelBasedSoft},
  title       = {A Fair Objective for Human-Empowerment-Preserving {AI}: Desiderata, Design, and Likely Behavioral Consequences},
  author      = {Heitzig, Jobst and Potham, Ram},
  date        = {2026-08-08},
  eprint      = {2608.08240v1},
  eprinttype  = {arXiv},
  eprintclass = {cs.AI},
  version     = {1},
  doi         = {10.48550/arXiv.2608.08240},
  url         = {https://arxiv.org/abs/2608.08240v1},
  urldate     = {2026-09-03},
  pubstate    = {prepublished},
}

@inproceedings{Jaques2019Influence,
  author    = {Jaques, Natasha and Lazaridou, Angeliki and Hughes, Edward and G{\"u}l{\c{c}}ehre, {\c{C}}a{\u{g}}lar and Ortega, Pedro A. and Strouse, DJ and Leibo, Joel Z. and de Freitas, Nando},
  title     = {Social Influence as Intrinsic Motivation for Multi-Agent Deep Reinforcement Learning},
  booktitle = {International Conference on Machine Learning},
  series    = {Proceedings of Machine Learning Research},
  volume    = {97},
  pages     = {3040--3049},
  year      = {2019}
}

@inproceedings{Jiang2021EOI,
  author    = {Jiang, Jiechuan and Lu, Zongqing},
  title     = {The Emergence of Individuality},
  booktitle = {International Conference on Machine Learning},
  series    = {Proceedings of Machine Learning Research},
  volume    = {139},
  pages     = {4992--5001},
  year      = {2021}
}

@inproceedings{Kim2023VM3AC,
  author    = {Kim, Woojun and Jung, Whiyoung and Cho, Myungsik and Sung, Youngchul},
  title     = {A Variational Approach to Mutual Information-Based Coordination for Multi-Agent Reinforcement Learning},
  booktitle = {International Conference on Autonomous Agents and Multiagent Systems (AAMAS)},
  pages     = {40--48},
  year      = {2023}
}

@inproceedings{Klyubin2005EmpowermentUniversal,
  title      = {Empowerment: A Universal Agent-Centric Measure of Control},
  shorttitle = {Empowerment},
  booktitle  = {2005 {{IEEE Congress}} on {{Evolutionary Computation}}},
  author     = {Klyubin, A.S. and Polani, D. and Nehaniv, C.L.},
  date       = {2005-09},
  volume     = {1},
  pages      = {128-135 Vol.1},
  issn       = {1941-0026},
  doi        = {10.1109/CEC.2005.1554676},
  eventtitle = {2005 {{IEEE Congress}} on {{Evolutionary Computation}}}
}

@article{Kulveit2025Gradual,
  title={Gradual Disempowerment: Systemic Existential Risks from Incremental AI Development},
  author={Kulveit, Jan and Douglas, Raymond and Ammann, Nora and Turan, Deger and Krueger, David and Duvenaud, David},
  journal={arXiv preprint arXiv:2501.16946},
  year={2025}
}

@article{KupfermanVardi2001Safety,
  title={Model Checking of Safety Properties},
  author={Kupferman, Orna and Vardi, Moshe Y.},
  journal={Formal Methods in System Design},
  volume={19},
  number={3},
  pages={291--314},
  year={2001},
  doi={10.1023/A:1011254632723}
}

@inproceedings{Li2021CDS,
  author    = {Li, Chenghao and Wang, Tonghan and Wu, Chengjie and Zhao, Qianchuan and Yang, Jun and Zhang, Chongjie},
  title     = {Celebrating Diversity in Shared Multi-Agent Reinforcement Learning},
  booktitle = {Advances in Neural Information Processing Systems},
  volume    = {34},
  year      = {2021}
}

@inproceedings{Mahajan2019MAVEN,
  author    = {Mahajan, Anuj and Rashid, Tabish and Samvelyan, Mikayel and Whiteson, Shimon},
  title     = {{MAVEN}: Multi-Agent Variational Exploration},
  booktitle = {Advances in Neural Information Processing Systems},
  volume    = {32},
  year      = {2019}
}

@inproceedings{massey1990causality,
  title={Causality, feedback and directed information},
  author={Massey, James L.},
  booktitle={Proc. Int. Symp. Inf. Theory Applic. (ISITA-90)},
  pages={303--305},
  year={1990}
}

@inproceedings{MannaPnueli1990Hierarchy,
  title={A Hierarchy of Temporal Properties},
  author={Manna, Zohar and Pnueli, Amir},
  booktitle={Proceedings of the Ninth Annual ACM Symposium on Principles of Distributed Computing (PODC)},
  year={1990},
  doi={10.1145/93385.93442}
}

@inproceedings{Mohamed2015VIM,
  title={Variational Information Maximisation for Intrinsically Motivated Reinforcement Learning},
  author={Mohamed, Shakir and Rezende, Danilo Jimenez},
  booktitle={Advances in Neural Information Processing Systems},
  volume={28},
  year={2015}
}

@inproceedings{Myers2024Assist,
  title={Learning to Assist Humans without Inferring Rewards},
  author={Myers, Vivek and Ellis, Evan and Levine, Sergey and Eysenbach, Benjamin and Dragan, Anca D.},
  booktitle={Advances in Neural Information Processing Systems},
  volume={37},
  year={2024},
  pages={71540--71567},
  doi={10.52202/079017-2285}
}

@inproceedings{Pong2020SkewFit,
  author    = {Pong, Vitchyr H. and Dalal, Murtaza and Lin, Steven and Nair, Ashvin and Bahl, Shikhar and Levine, Sergey},
  title     = {Skew-Fit: State-Covering Self-Supervised Reinforcement Learning},
  booktitle = {International Conference on Machine Learning},
  series    = {Proceedings of Machine Learning Research},
  volume    = {119},
  pages     = {7783--7792},
  year      = {2020}
}

@inproceedings{Park2024METRA,
  author    = {Park, Seohong and Rybkin, Oleh and Levine, Sergey},
  title     = {{METRA}: Scalable Unsupervised {RL} with Metric-Aware Abstraction},
  booktitle = {International Conference on Learning Representations},
  year      = {2024}
}

@inproceedings{Reddy2022FirstContact,
  title={First Contact: Unsupervised Human-Machine Co-Adaptation via Mutual Information Maximization},
  author={Reddy, Siddharth and Levine, Sergey and Dragan, Anca},
  booktitle={Advances in Neural Information Processing Systems},
  volume={35},
  year={2022}
}

@inproceedings{Shi2024Opacity,
  author    = {Shi, Chongyang and Bu, Yuheng and Fu, Jie},
  title     = {Information-Theoretic Opacity-Enforcement in Markov Decision Processes},
  booktitle = {Proceedings of the Thirty-Third International Joint Conference on Artificial Intelligence (IJCAI)},
  year      = {2024}
}

@incollection{Salge2014Empowerment,
  title={Empowerment---An Introduction},
  author={Salge, Christoph and Glackin, Cornelius and Polani, Daniel},
  booktitle={Guided Self-Organization: Inception},
  editor={Prokopenko, Mikhail},
  pages={67--114},
  doi={10.1007/978-3-642-53734-9_4},
  year={2014},
  publisher={Springer}
}

@inproceedings{Sharma2020DADS,
  title={Dynamics-Aware Unsupervised Discovery of Skills},
  author={Sharma, Archit and Gu, Shixiang and Levine, Sergey and Kumar, Vikash and Hausman, Karol},
  booktitle={International Conference on Learning Representations},
  year={2020}
}

@inproceedings{Shi2026IMAS,
  author    = {Shi, Chongyang and Suttle, Wesley A. and Dorothy, Michael and Fu, Jie},
  title     = {{IMAS}$^2$: Joint Agent Selection and Information-Theoretic Coordinated Perception in {Dec-POMDPs}},
  booktitle = {International Conference on Autonomous Agents and Multiagent Systems (AAMAS)},
  year      = {2026},
  doi       = {10.65109/STZK9664}
}

@article{Salge2017ThreeLaws,
  title={Empowerment as Replacement for the Three Laws of Robotics},
  author={Salge, Christoph and Polani, Daniel},
  journal={Frontiers in Robotics and AI},
  volume={4},
  pages={25},
  year={2017}
}

@inproceedings{Strouse2018Intentions,
  title={Learning to Share and Hide Intentions using Information Regularization},
  author={Strouse, Daniel J. and Kleiman-Weiner, Max and Tenenbaum, Josh and Botvinick, Matthew M. and Schwab, David J.},
  booktitle={Advances in Neural Information Processing Systems},
  volume={31},
  year={2018}
}

@article{ToroIcarte2022RewardMachines,
  title={Reward Machines: Exploiting Reward Function Structure in Reinforcement Learning},
  author={Toro Icarte, Rodrigo and Klassen, Toryn Q. and Valenzano, Richard and McIlraith, Sheila A.},
  journal={Journal of Artificial Intelligence Research},
  volume={73},
  pages={173--208},
  year={2022},
  doi={10.1613/jair.1.12440}
}

@article{Tiomkin2025ProcessEmpowerment,
  title = {Process Empowerment for Robust Intrinsic Motivation},
  author = {Tiomkin, Stas and Salge, Christoph and Polani, Daniel},
  date = {2025-08},
  journaltitle = {Journal of Physics: Complexity},
  shortjournal = {J. Phys. Complex.},
  volume = {6},
  number = {3},
  pages = {035011},
  publisher = {IOP Publishing},
  issn = {2632-072X},
  doi = {10.1088/2632-072X/adf2ec},
  url = {https://doi.org/10.1088/2632-072X/adf2ec},
  urldate = {2026-06-19},
  langid = {english}
}

@article{Tiomkin2017Unified,
  title={A Unified Bellman Equation for Causal Information and Value in Markov Decision Processes},
  author={Tiomkin, Stas and Tishby, Naftali},
  journal={arXiv preprint arXiv:1703.01585},
  year={2017}
}

@inproceedings{Turner2021Power,
  title={Optimal Policies Tend to Seek Power},
  author={Turner, Alexander Matt and Smith, Logan and Shah, Rohin and Critch, Andrew and Tadepalli, Prasad},
  booktitle={Advances in Neural Information Processing Systems},
  volume={34},
  year={2021}
}

@inproceedings{Wang2020ROMA,
  author    = {Wang, Tonghan and Dong, Heng and Lesser, Victor and Zhang, Chongjie},
  title     = {{ROMA}: Multi-Agent Reinforcement Learning with Emergent Roles},
  booktitle = {International Conference on Machine Learning},
  series    = {Proceedings of Machine Learning Research},
  volume    = {119},
  pages     = {9876--9886},
  year      = {2020}
}

@inproceedings{Yang2023HMASD,
  author    = {Yang, Mingyu and Yang, Yaodong and Lu, Zhenbo and Zhou, Wengang and Li, Houqiang},
  title     = {Hierarchical Multi-Agent Skill Discovery},
  booktitle = {Advances in Neural Information Processing Systems},
  volume    = {36},
  year      = {2023}
}

@online{Yang2025Bystander,
  author     = {Yang, Claire and Zhang, Claire Jie and Cakmak, Maya and Kleiman-Weiner, Max},
  title      = {When Assisting One Disempowers Another},
  year       = {2026},
  eprint     = {2511.04177v2},
  eprinttype = {arxiv},
  version    = {2},
  doi        = {10.48550/arXiv.2511.04177},
  note       = {Preprint; previously titled ``When Empowerment Disempowers''}
}

@inproceedings{Yang2022LTLIntractable,
  title={On the (In)Tractability of Reinforcement Learning for {LTL} Objectives},
  author={Yang, Cambridge and Littman, Michael L. and Carbin, Michael},
  booktitle={Proceedings of the Thirty-First International Joint Conference on Artificial Intelligence (IJCAI)},
  year={2022}
}

\appendix
\section{Proofs}
\label{app:proofs}

\subsection{Proof of Theorem~\ref{thm:bridge}}

\paragraph{Upper inequality (feasibility).} The human's realised behaviour with the goal marginalised out is the input distribution
\begin{equation}
    p_{\pih}(\mathbf{a}_h \parallel \traj_{0:T-1}) = \prod_{t=1}^T p_{\pih}(a_{h,t} \mid \mathbf{a}_{h,1:t-1}, \traj_{0:t-1})
    = \prod_{t=1}^T \sum_{g} \pih(a_{h,t} \mid s_{t-1}, g)\; p(g \mid \mathbf{a}_{h,1:t-1}, \traj_{0:t-1}).
    \label{eq:realised-input-law}
\end{equation}
The mixture has memory in the actions even when each per-goal policy is Markov, because past actions carry evidence about which $g$ is in play; the conditional $p(g \mid \cdots)$ is bookkeeping from the marginalisation, not any agent's belief. Equation~\eqref{eq:realised-input-law} is causally conditioned (each factor reads only the past), so it lies in the feasible set of the maximisation defining $\empow(s \mid \pir)$ against the same channel, and $\mathbb{I}_{\pih,\pir}(\mathbf{A}_h \to \traj \mid s) \le \empow(s)$.

\paragraph{Middle inequality (Massey's converse).} Identify $G$ as the message and $\mathbf{A}_h$ as the causal channel inputs. The induced kernel~\eqref{eq:induced-channel} is the conditional law of $S_t$ given any history containing $(S_{t-1}, A_{h,t})$ and does not read the goal, so
\begin{equation}
    \mathbb{I}(G; S_t \mid \mathbf{A}_{h,1:t}, \traj_{0:t-1}, s) = 0 \qquad \text{for every } t:
    \label{eq:message-independent-kernel}
\end{equation}
adjoining the message to the input changes nothing. Summing the chain rule $\mathbb{I}(G, \mathbf{A}_{h,1:t}; S_t \mid \traj_{0:t-1}, s) = \mathbb{I}(\mathbf{A}_{h,1:t}; S_t \mid \traj_{0:t-1}, s) + \mathbb{I}(G; S_t \mid \mathbf{A}_{h,1:t}, \traj_{0:t-1}, s)$ over $t$ gives $\mathbb{I}(G, \mathbf{A}_h \to \traj \mid s) = \mathbb{I}(\mathbf{A}_h \to \traj \mid s)$. Expanding the left side by the chain rule in the other order, and using $\sum_{t} \mathbb{I}(G; S_t \mid \traj_{0:t-1}, s) = \mathbb{I}(G; \traj \mid s)$ (the chain rule for mutual information, $G$ having no time index), yields the identity~\eqref{eq:massey-identity}; nonnegativity of its second term is the inequality \cite{massey1990causality}.

The detour through directed information is forced. $G \to \mathbf{A}_h \to \traj$ is not a Markov chain under feedback, so the undirected data-processing inequality is unavailable: the action is a common effect of the goal and the realised states, and conditioning on it couples them. A two-line counterexample: let the dynamics be pure noise (states independent of actions) and let the human's single action report whether the realised first state matches their goal. Then $\mathbb{I}(G; \traj \mid s) = 0$, yet given the action the two are dependent, $\mathbb{I}(G; \traj \mid \mathbf{A}_h, s) > 0$, so $G \perp \traj \mid \mathbf{A}_h$ fails.

\paragraph{Lower inequality (Fano floor).} Define the satisfaction decoder $\Ghat(\traj)$: output $g$ if $\traj \in \Trajs_g$, and a fixed default $\gnull \in \Gh$ if $\traj$ satisfies no goal. By disjointness (Assumption~\ref{ass:atomic-goals}) at most one goal is satisfied, so $\Ghat$ is single-valued and everywhere defined; the default only raises the correct-guess probability for $g = \gnull$, and no goal is certified at time zero, so no trivial immediate decode occurs. Given $G = g$, a correct guess is at least as probable as satisfaction, which the capability lower-bounds pointwise, since $\gamma_h \le 1$ gives $\gamma_h^{\tau_g}\mathbf{1}(\traj \in \Trajs_g) \le \mathbf{1}(\traj \in \Trajs_g)$:
\begin{equation}
    \mathbb{P}(\Ghat = g \mid G = g, s) \ge \mathbb{P}(\traj \in \Trajs_g \mid G = g, s) \ge \Ch(s, g \mid \pig).
\end{equation}
Averaging over the uniform prior and using $x^\zeta \le x$ on $[0,1]$ for $\zeta \ge 1$,
\begin{equation}
    \mathbb{P}(\Ghat = G \mid s) \ge \frac{1}{\Ngoals}\sum_{g} \Ch(s, g \mid \pig) \ge \frac{1}{\Ngoals}\sum_{g} \Ch(s, g \mid \pig)^{\zeta} = 2^{\,I_h(s) - \log_2 \Ngoals},
\end{equation}
so the decoding error obeys $P_e(s) \le P_e^{\max}(s) = 1 - 2^{I_h(s) - \log_2 \Ngoals}$. Because $\Ghat$ is a function of $\traj$, the data-processing inequality (valid here: $G \to \traj \to \Ghat$ is a Markov chain) gives $\mathbb{H}(G \mid \traj, s) \le \mathbb{H}(G \mid \Ghat, s)$, and Fano's inequality bounds the latter by $f(P_e(s))$ with $f(x) = h_2(x) + x\log_2(\Ngoals - 1)$. The function $f$ increases up to its peak at the random-guessing rate $(\Ngoals-1)/\Ngoals$, where $f\big(\tfrac{\Ngoals-1}{\Ngoals}\big) = \log_2 \Ngoals$, so
\begin{equation}
    \mathbb{H}(G \mid \traj, s) \le f\Big(\min\big(P_e^{\max}(s), \tfrac{\Ngoals-1}{\Ngoals}\big)\Big)
\end{equation}
holds in both regimes: for $P_e^{\max}(s)$ below the peak by monotonicity, since $P_e(s) \le P_e^{\max}(s)$, and above it because the right side equals $\log_2 \Ngoals$, which bounds the equivocation always. Substituting into $\Rev(s) = \log_2 \Ngoals - \mathbb{H}(G \mid \traj, s)$ gives~\eqref{eq:fano-floor}. Every step uses only Assumptions~\ref{ass:goal-prior}--\ref{ass:atomic-goals}, disjointness, and the pointwise bound, so the floor holds for any disjoint family of trajectory goals under the realised (not optimal) policy. \hfill$\square$

\subsection{Proof of Lemma~\ref{lem:monotone-floor}, and a closed form}

$P_e^{\max}(s)$ is strictly decreasing in $I_h(s)$, and $f$ is strictly increasing on $[0, (\Ngoals-1)/\Ngoals]$, since $f'(x) = \log_2 \frac{(\Ngoals-1)(1-x)}{x} \ge 0$ if and only if $x \le \frac{\Ngoals-1}{\Ngoals}$. For $I_h(s) \ge 0$ the clamp is inactive, so $\Bs = \log_2 \Ngoals - f(P_e^{\max}(s))$ is strictly increasing in $I_h(s)$. As $I_h(s) \to \log_2 \Ngoals$ we get $P_e^{\max}(s) \to 0$, hence $f \to 0$ and $\Bs \to \log_2 \Ngoals$. \hfill$\square$

Substituting $P_e^{\max}$ into $f$ and simplifying with $c = 2^{I_h(s)}/\Ngoals$ gives the closed form, for $I_h(s) \in [0, \log_2 \Ngoals]$:
\begin{equation}
    \Bs = \frac{2^{I_h(s)}}{\Ngoals}\, I_h(s) + \Big(1 - \frac{2^{I_h(s)}}{\Ngoals}\Big) \log_2 \frac{\Ngoals - 2^{I_h(s)}}{\Ngoals - 1}.
    \label{eq:fano-closed-form}
\end{equation}

\subsection{Proof of Remark~\ref{rem:deterministic-pursuit}}

By induction on $t$: $A_{h,1}$ is a deterministic function of $(G, S_0)$, and if $\mathbf{A}_{h,1:t-1}$ is a deterministic function of $(G, \traj_{0:t-2})$, then $A_{h,t} = \pih(\cdot \mid \text{history}, G)$ is a deterministic function of $(G, \traj_{0:t-1})$. Hence each term of $\mathbb{I}(\mathbf{A}_h \to \traj \mid G, s) = \sum_t \mathbb{I}(\mathbf{A}_{h,1:t}; S_t \mid \traj_{0:t-1}, G, s)$ conditions on variables that determine $\mathbf{A}_{h,1:t}$, and vanishes. The claim then follows from the identity~\eqref{eq:massey-identity}. \hfill$\square$

\subsection{Proof of Proposition~\ref{prop:z-decomposition}}

First, the goal-conditioned trajectory law depends on $g$ only through its class. Let $g \sim_s g'$. The law of the history under $g$ is a product of kernel factors and policy factors $\pih(\cdot \mid \eta, g)$ evaluated at realised histories $\eta$; by induction on $t$, every history with positive probability under $g$ lies in $\mathcal{H}_s$, where the two policies agree, so the laws of all histories, and of $\traj$, coincide under $g$ and $g'$. Hence $p(\traj \mid g, s) = p(\traj \mid \Zpol = [g], s)$ is well defined, $G \perp \traj \mid \Zpol$, and since $\Zpol$ is a function of $G$, $G \to \Zpol \to \traj$ is a Markov chain.

For the decomposition: $\Zpol = f(G)$ gives $\mathbb{H}(G \mid \traj, s) = \mathbb{H}(G, \Zpol \mid \traj, s) = \mathbb{H}(\Zpol \mid \traj, s) + \mathbb{H}(G \mid \Zpol, \traj, s)$, and $G \perp \traj \mid \Zpol$ reduces the last term to $\mathbb{H}(G \mid \Zpol, s)$. Likewise $\mathbb{H}(G \mid s) = \mathbb{H}(\Zpol \mid s) + \mathbb{H}(G \mid \Zpol, s)$. Subtracting,
$\Rev(s) = \mathbb{H}(\Zpol \mid s) - \mathbb{H}(\Zpol \mid \traj, s) = \mathbb{I}(\Zpol; \traj \mid s) \le \mathbb{H}(\Zpol \mid s)$. \hfill$\square$

\subsection{Proof of Proposition~\ref{prop:capacity-separation}}

\textit{Construction.} States are pairs $(x, \mathbf{b})$ with location $x \in \{0, 1, \dots, \Ngoals\}$ ($0$ a hub; each $j \ge 1$ an absorbing room whose target set $\mathcal{S}_{g_j} = \{(j, \mathbf{b}) : \mathbf{b} \in \{0,1\}^m\}$ induces the first-outcome goal $g_j$) and board configuration $\mathbf{b} \in \{0,1\}^m$. The initial state is $s = (0, \mathbf{0})$, in no target set. The human's actions are $\{\textsf{go}_1, \dots, \textsf{go}_{\Ngoals}\} \cup \{\textsf{flip}_1, \dots, \textsf{flip}_m\} \cup \{\textsf{wait}\}$. Under $\pir^{\mathrm{pix}}$, $\textsf{flip}_i$ toggles $b_i$ and every $\textsf{go}_j$ is inert; under $\pir^{\mathrm{ass}}$, $\textsf{go}_j$ moves $0 \to j$, absorbing thereafter (windows padded per Section~\ref{sec:setup}), and every $\textsf{flip}_i$ is inert. Pursuit is $\pih(\textsf{go}_j \mid s, g_j) = 1$.

(1) Under $\pir^{\mathrm{pix}}$ the induced channel is deterministic. The distinguishable action classes per step are the $m+1$ classes $\{\textsf{flip}_1, \dots, \textsf{flip}_m, \textsf{wait}\}$ (every $\textsf{go}_j$ coincides with $\textsf{wait}$), and distinct class sequences yield distinct trajectories, the class at step $t$ being recoverable from $(S_{t-1}, S_t)$. Hence $\empow(s) \le \mathbb{H}(\traj \mid s) \le T\log_2(m+1)$, attained by the uniform open-loop input over class sequences, so $\empow(s) = T\log_2(m+1)$. Under pursuit every chosen action is inert, so $\traj$ is one constant trajectory for every $G$: $\Rev(s) = 0$, and no target is reachable, so every capability is $0$.

(2) Under $\pir^{\mathrm{ass}}$ the achievable trajectories are ``wait $k-1$ steps, then enter room $j$'' for $(j,k) \in \Gh \times \{1, \dots, T\}$, plus the all-wait trajectory: $\Ngoals T + 1$ in total, through a deterministic channel injective on distinguishable inputs, so $\empow(s) = \log_2(\Ngoals T + 1)$. Pursuit enters room $j$ at $k = 1$ with certainty, so every capability is $1$ at $\gamma_h = 1$, and $\traj$ determines $G$: $\Rev(s) = \log_2 \Ngoals$.

\textit{Threshold.} For $m \ge \Ngoals + 1$, Bernoulli's inequality gives $(m+1)^T \ge (\Ngoals + 2)^T \ge 1 + T(\Ngoals + 1) > \Ngoals T + 1$. \hfill$\square$

\subsection{Proof of Proposition~\ref{prop:scramble-separation}}

(1) Under the uniform prior, scrambling replaces the channel $g \mapsto p(\cdot \mid g, s)$ by $g \mapsto p(\cdot \mid \sigma(g), s)$. The trajectory marginal is unchanged, $\frac{1}{\Ngoals}\sum_g p(\traj \mid \sigma(g), s) = \frac{1}{\Ngoals}\sum_{g'} p(\traj \mid g', s)$, and the conditional entropy is a uniform average over the same set of distributions, by the substitution $g' = \sigma(g)$; hence both terms of $\mathbb{I}(G;\traj \mid s) = \mathbb{H}(\traj \mid s) - \mathbb{H}(\traj \mid G, s)$ are unchanged.

(2) With goal $g$, the scrambled trajectory is distributed as an original trajectory with goal $\sigma(g)$. If $\sigma(g) \ne g$, the hypothesis applied to the original law for $\sigma(g)$, with $g' = g$, gives $\mathbb{P}(\traj \in \Trajs_g) = 0$ under the scrambled law, and the certificate-time expectation~\eqref{eq:capability-def} vanishes with it; if $\sigma(g) = g$ the law, and the capability, are unchanged. For fixed-point-free $\sigma$ every capability is $0$, the sum inside $I_h$ is $0$, $I_h(s) \to -\infty$, and $\Bs = 0$ by the clamp.

The final claim takes the configuration of Table~\ref{tab:scramble-instance} and its swap: identical $\Rev(s)$, capability profiles $(c, c)$ and $(0,0)$. \hfill$\square$

\subsection{Proof of Theorem~\ref{thm:vim}}

Revelation is the relative entropy between the joint law of $(G, \traj)$ and the product of its marginals, $\Rev(s) = D\big(p(g, \traj \mid s) \,\|\, p(g \mid s)\,p(\traj \mid s)\big)$. Relative entropy cannot increase when the same measurable map is applied to both arguments. Apply the map $(g, \traj) \mapsto \mathbf{1}(\traj \in \Trajs_g)$. Under the joint law its image is Bernoulli with parameter $\mathbb{P}(\traj \in \Trajs_G \mid s) = \Pbar(s)$. Under the product law, with $G$ uniform and independent of $\traj$, it is Bernoulli with parameter $\mathbb{E}_{\traj}\big[\tfrac{1}{\Ngoals}\sum_g \mathbf{1}(\traj \in \Trajs_g)\big] = \bar K(s)/\Ngoals$. The relative entropy between two Bernoulli laws is $d$, which is \eqref{eq:vim-bound}. If $\bar K/\Ngoals \in \{0, 1\}$ then every reachable trajectory satisfies no goal, or every goal, so $\Pbar$ takes the same value and $d = 0$; otherwise both arguments lie in $(0, 1)$. Expanding $d$ and collecting the $\log_2 \Ngoals$ terms gives \eqref{eq:vim-expanded}. \hfill$\square$

\paragraph{The decoder route (Remark~\ref{rem:vim-decoder}).} With the uniform prior, choose the decoder that reads $\traj$ through its satisfaction indicators, a softmax with one log-odds weight $W$:
\begin{equation}
    q(g \mid \traj) = \frac{\exp\big(W\,\mathbf{1}(\traj \in \Trajs_g)\big)}{\sum_{g'} \exp\big(W\,\mathbf{1}(\traj \in \Trajs_{g'})\big)}.
\end{equation}
Since each indicator is $0$ or $1$, the partition sum is $\Ngoals + (e^W - 1)K(\traj)$, where $K(\traj)$ counts the goals $\traj$ satisfies. Substituting into~\eqref{eq:vim-generic},
\begin{equation}
    \Rev(s) \ge \log_2 \Ngoals + W\,\Pbar(s)\log_2 e - \mathbb{E}\big[\log_2\big(\Ngoals + (e^W - 1)K(\traj)\big)\big],
\end{equation}
using $\mathbb{E}[\mathbf{1}(\traj \in \Trajs_G)] = \Pbar(s)$. Jensen's inequality on the concave $\log$ bounds the last term by $\log_2(\Ngoals + (e^W - 1)\bar K(s))$, the argument being positive for every finite $W$. Optimising the resulting bound over $W$ gives
\begin{equation}
    e^{W^{*}} = \frac{\Pbar(s)\,(\Ngoals - \bar K(s))}{\bar K(s)\,(1 - \Pbar(s))},
\end{equation}
and substituting $W^{*}$ back yields \eqref{eq:vim-expanded}, the expanded form of \eqref{eq:vim-bound}. The zero-revelation case $\Pbar = \bar K/\Ngoals$ makes $e^{W^*} = 1$, and $W^* > 0$ exactly when $\Pbar > \bar K/\Ngoals$.

\end{document}